\pdfoutput=1

\documentclass[11pt]{article}

\usepackage[final]{acl}

\usepackage{longtable}
\usepackage{times}
\usepackage{latexsym}
\usepackage{bbm}
\usepackage{graphicx, multicol}

\usepackage{amsmath}
\usepackage{cleveref}
\crefname{figure}{Figure}{Figures}
\crefname{table}{Table}{Tables}
\crefname{appendix}{Appendix}{Appendices}
\crefname{section}{Section}{Sections}
\crefname{equation}{Eq.}{Eqs.}

\usepackage{ulem}

\usepackage{textcomp}  
\usepackage{scalerel}  

\def\blueberries{\scalerel*{\includegraphics{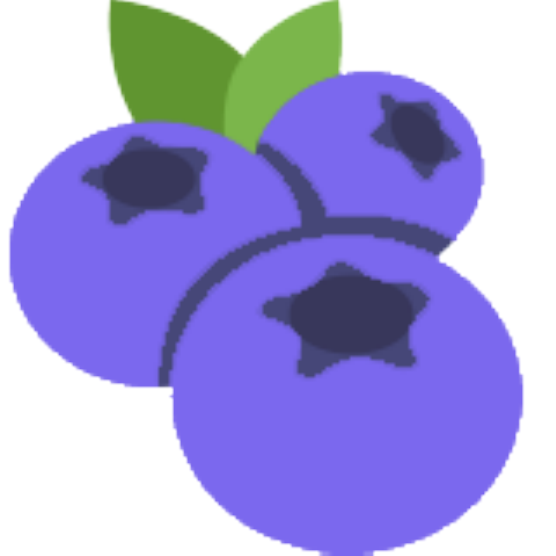}}{\textrm{\textbigcircle}}}
\def\lemon{\scalerel*{\includegraphics{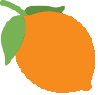}}{\textrm{\textbigcircle}}}

\usepackage[T1]{fontenc}

\usepackage[utf8]{inputenc}

\usepackage{microtype}

\usepackage{booktabs} 
\usepackage{array}
\usepackage{bm}
\usepackage{amsmath}
\usepackage{amsfonts}
\usepackage{amssymb} 
 
\usepackage{enumitem} 

\usepackage[T1]{fontenc}

\usepackage{color}
\usepackage{soul}
\usepackage[textsize=scriptsize]{todonotes}
\definecolor{pigment}{rgb}{0.2, 0.2, 0.6}

\definecolor{blue}{RGB}{0, 93, 170}			
\definecolor{darkgreen}{HTML}{3bb35b}

\newcommand{\indep}{\perp \!\!\! \perp}

\title{
Slangvolution: A Causal Analysis of Semantic Change \\ and Frequency Dynamics in Slang 
}

\author{Daphna Keidar\textsuperscript{\blueberries,}\thanks{\; Equal contribution.} ~\;~ 
Andreas Opedal\textsuperscript{ \blueberries,}\footnotemark[1] ~\;~
Zhijing Jin\textsuperscript{\lemon, \blueberries} ~\;~
Mrinmaya Sachan\textsuperscript{\blueberries} \\

\textsuperscript{\blueberries}ETH Zürich, ~\;~\;~ 
\textsuperscript{\lemon}Max Planck Institute for Intelligent Systems, Tübingen, Germany \\ 
\href{mailto:dkeidar@ethz.ch}{\texttt{dkeidar@ethz.ch}},~\;~  \href{mailto:andreas.opedal@inf.ethz.ch}{\texttt{andreas.opedal@inf.ethz.ch}},\\
\href{mailto:zjin@tue.mpg.de}{\texttt{zjin@tue.mpg.de}},~\;~  \href{mailto:mrinmaya.sachan@inf.ethz.ch}{\texttt{mrinmaya.sachan@inf.ethz.ch}}}

\begin{document}
\maketitle
\begin{abstract}
Languages are continuously undergoing changes, and the  mechanisms that underlie these changes are still a matter of debate. In this work, we approach language evolution through the lens of causality in order to model not only how various distributional factors associate with language change, but how they causally affect it. In particular, we study slang, which is an informal language that is typically restricted to a specific group or social setting. We analyze the semantic change and frequency shift of slang words and compare them to those of standard, nonslang words. With causal discovery and causal inference techniques, we measure the effect that word type (slang/nonslang) has on both semantic change and frequency shift, as well as its relationship to frequency, polysemy and part of speech. Our analysis provides some new insights in the study of language change, e.g., we show that slang words undergo less semantic change but tend to have larger frequency shifts over time.\footnote{Our code, along with the data, is made available at \href{https://github.com/andreasopedal/slangvolution}{https://github.com/andreasopedal/slangvolution}.}
\end{abstract}

\section{Introduction}
Language is a continuously evolving system, constantly resculptured by its speakers. 
The forces that drive this evolution are many, ranging from phonetic convenience to sociocultural changes \cite{blank-1999}. 
In particular, the meanings of words and the
frequencies in which they are used 
are not static, but rather evolve over time. 
Several previous works, in both historical 
and computational linguistics, have described diachronic mechanisms, often suggesting causal relationships. 
For example, semantic change, i.e. change in the meaning of a word, has both been suggested to \textit{cause} \cite{wilkins_polysemy_semantic_change, hopper_grammaticalization} and \textit{be caused by} \cite{hamilton-etal-2016-diachronic} polysemy, while also part of speech (POS) has been implied to be a causal factor behind semantic change \cite{Dubossarsky2016VerbsCM}.
However, none of these studies perform a causal analysis to verify these claims. Causality \cite{pearl2009causal} allows us to not only infer causal effects between pairs of variables, but also model their interactions with other related factors.
\begin{figure}[t]
    \centering
    \includegraphics[width=0.48\textwidth]{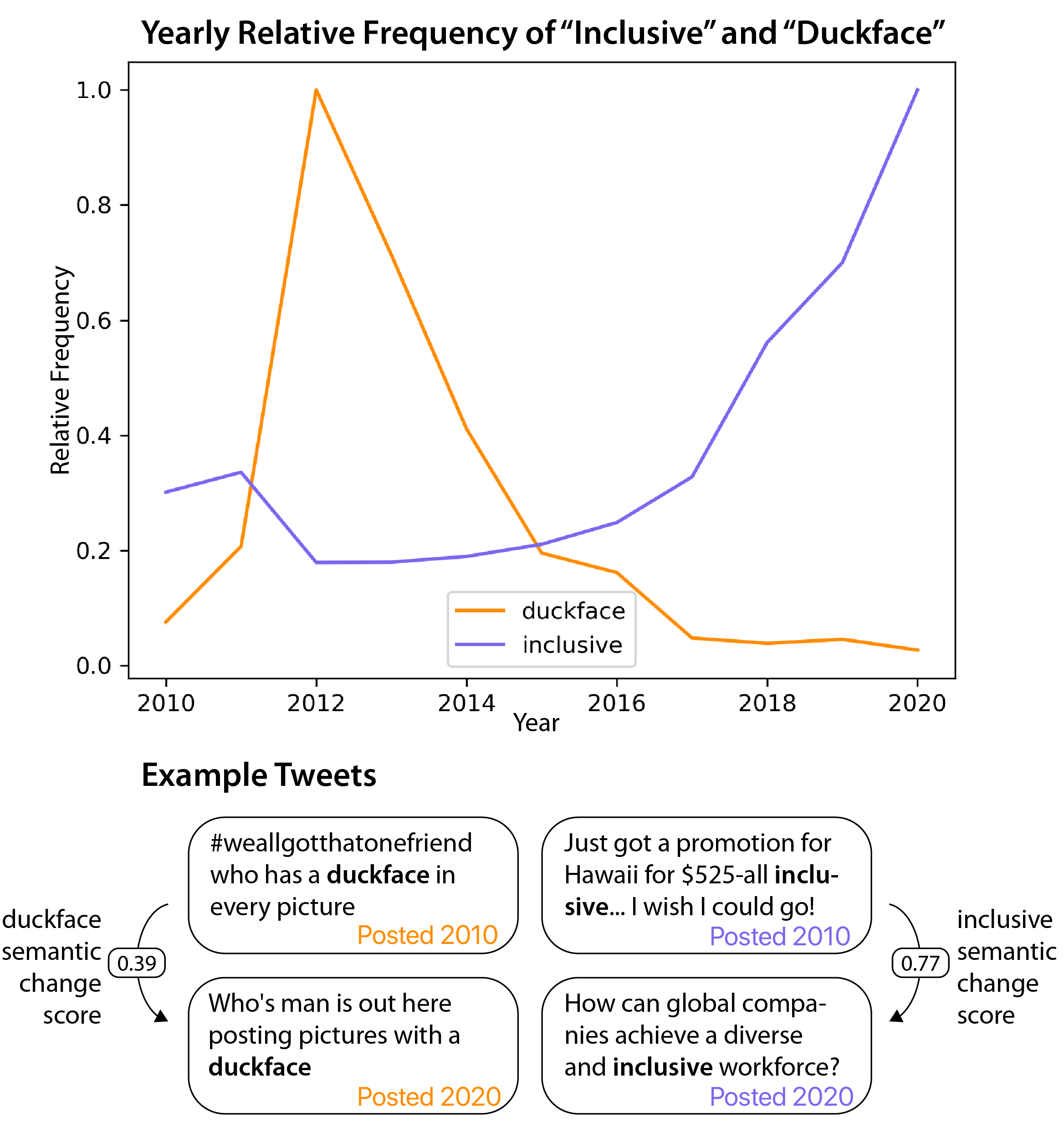}
    \caption{We observe very different change dynamics for the slang word ``duckface'' and the nonslang word ``inclusive.'' ``Inclusive'' has acquired a new meaning, reflected in a high semantic change score of $0.77$ as measured by our model. ``Duckface'' undergoes little semantic change, scored $0.39$ by our model, while its usage frequency varies greatly. 
    }
    \label{fig:example}
\end{figure}

In this work, we focus on the linguistic evolution of slang, 
defined as colloquial and informal language commonly associated with particular groups \cite{ingroup_language_college_students, black_slang}, 
and use a causal framework to compare the change dynamics of slang words to those of standard language. 
More specifically, we compare the \textit{semantic change} as well as the changes in frequency, i.e., \textit{frequency shift}, over time between slang words and standard, nonslang words. We learn a \textit{causal graphical model} \cite{spirtes2000causation} to assess how these variables interact with other factors they have been previously found to correlate with, 
such as \textit{frequency}, \textit{polysemy} and \textit{part of speech} \cite{Dubossarsky2016VerbsCM, hamilton-etal-2016-diachronic}. Having discovered a graph, we proceed to use \textit{do-calculus} \cite{pearl-do-calculus} to evaluate the causal effects of a word's \textit{type} (slang/nonslang) on semantic change and frequency shift. 
\newline\indent
Semantic change is measured using 
the average pairwise distance (APD) \cite{sagi-etal-2009-semantic, giulianelli-etal-2020-analysing} between time-separated contextualized representations, which were obtained from a Twitter corpus via a bi-directional language model \cite{roberta}. Our method builds on recent semantic change literature \cite{schlechtweg2020semeval1}, with novel additions of dimensionality reduction and a combined distance function.  
\newline\indent
By 
deploying a causal analysis, we establish that there is not just an association, but a direct effect of a word's type
on its semantic change and frequency shift.
We find that a word being \textit{slang} causes it to undergo slower semantic change and more rapid decreases in frequency. 
To illustrate, consider the slang word ``duckface'' and the nonslang word ``inclusive'' as shown in \cref{fig:example}. Duckface is a face pose commonly made for photos \cite{duckface_nytimes} in the early 2010s, and while it has largely decreased in frequency since, its meaning has not changed.
In contrast, the nonslang word ``inclusive'' has developed a new usage in recent years
\cite{merriam-webster-inclusive} and was given a high semantic change score by our model.
\newline\indent
Our analysis also sheds light on a couple of previous findings in the diachronic linguistics literature. We find support for the S-curve theory \cite{kroch_1989_scurve}, showing a causal effect from a word's polysemy to its frequency. This relationship is evident in the increase in frequency that the word ``inclusive'' displays in \cref{fig:example} after it develops a new meaning \cite{merriam-webster-inclusive}.
However, similar to \newcite{dubossarsky-etal-2017-outta}, we do not find causal links to semantic change from frequency, polysemy, or POS, which have been suggested in previous works \cite{hamilton-etal-2016-diachronic,Dubossarsky2016VerbsCM}.
\newline\indent
In summary, our main contributions are threefold: (i) we 
formalize the analysis of change dynamics in language with a causal framework;
(ii) we propose a semantic change metric that builds upon contextualized word representations; and (iii) we discover interesting insights about slang words and semantic change -- e.g., showing that the change dynamics of slang words are different from those of nonslang words, with slang words exhibiting both more rapid frequency fluctuations and less semantic change.
\section{Related Work}
\subsection{Semantic Change}
A typical method for measuring semantic change is by comparing word representations 
across time periods \cite{gulordava-baroni-2011-distributional,kim-etal-2014-temporal,jatowt_vis_change,kulkarni_linguistic_change, eger-mehler-2016-linearity, schlechtweg-etal-2019-wind}. 
With this approach, previous research has proposed laws relating semantic change to other linguistic properties \cite{Dubossarsky2015ABU, Xu_Kemp_laws, Dubossarsky2016VerbsCM, hamilton-etal-2016-diachronic}. For instance, \citet{Dubossarsky2016VerbsCM} find that verbs change faster than nouns,  
whereas \citet{hamilton-etal-2016-diachronic} 
discover that polysemous words change at a faster rate, while frequent words change slower.
However, the validity of some of these results has been questioned via case-control matching \cite{dubossarsky-etal-2017-outta}, highlighting the influence of word frequency on the representations and thus on the semantic change metric \cite{hellrich-hahn-2016-bad}. Such analyses can indeed give stronger evidence for causal effects. In this work we take a methodologically different approach, considering observational data alone for our causal analysis. \newline\indent
The aforementioned works rely on fixed word representations, 
whereas more recent approaches \cite{hu-etal-2019-diachronic, giulianelli-etal-2020-analysing} 
have proposed 
semantic change measures based on contextualized word embeddings \cite{ELMo, bert}, which can flexibly capture contextual nuances in word meaning. This has lead to a further stream of work on semantic change detection with contextualized embeddings \cite{capturing_evolution_more_clusters, semeval_kutuzov,schlechtweg2020semeval1,montariol-etal-2021-scalable,giulianelli2021grammatical,laicher-etal-2021-explaining}. We build upon this line of work and extend them using principal component analysis (PCA) and a combination of distance metrics. 

\subsection{
Characterization and Properties of Slang 
}
Slang is an informal, unconventional part of the language, often used in connection to a certain setting or societal trend
\cite{is_slang_a_word4_linguists}. 
It can reflect and establish a sense of belonging to a group \citep{ingroup_language_college_students, black_slang, peoples_poetry} or to a generation
\citep{slang-generation-gap, slang_teens_vs_adults, slang_age_based_patterns}.

\citet{Mattiello2005ThePO} highlights the 
role slang plays in enriching the language with neologisms, and claims that it follows unique word formation processes.
Inspired by this, \citet{slang_word_formation} propose a data-driven model for emulating the generation process of slang words that \citet{Mattiello2005ThePO} describes. Other computational works have focused on generating novel usages for existing slang words \cite{sun_cogsci19_slang, sun2021}, detecting and identifying slang as it occurs naturally in text \cite{pei-etal-2019-slang} or automatically giving explanatations for newly emerged slang \cite{ni-wang-2017-learning}.

Although the ephemerality of slang words has been described by studies from linguistics
\cite{ingroup_language_college_students, peoples_poetry},
this property has to the best of our knowledge not been previously verified by computational approaches. 

\section{Causal Methodology for Change Dynamics}
Examining change dynamics through a causal lens helps determine the existence of direct causal effects, by modeling the interactions between variables. For example, it allows us to conclude whether word type directly influences semantic change, or rather influences polysemy, which in turn causes semantic change.
In this section, we first give a short overview of relevant work on causality, before presenting how we apply these concepts to word change dynamics.
\subsection{Overview of Causal Discovery and Causal Inference}

A common framework for causal reasoning is through 
\textit{causal directed acyclic graphs} (DAGs) \cite{pearl2009causal}. A causal DAG consists of a pair $(G,P)$ where $G=(V,E)$ is a DAG and $P$ is a probability distribution over a set of variables. Each variable is represented by a node $v \in V$,
and the graph's edges $e \in E$ reflect causal relationships. 
There are two main tasks in causality. \textit{Causal discovery} 
is the task of uncovering the causal DAG that explains observed data. Assuming a causal DAG, the task of \textit{causal inference} then concerns determining the effect that intervening on a variable, often referred to as \textit{treatment}, will have on another variable, often referred to as \textit{outcome}.

The causal DAG is often inferred from domain knowledge or intuition. However, in cases where we cannot safely assume a known causal structure, causal discovery methods come in useful.
Constraint-based methods \cite{spirtes2000causation} form one of the main categories of causal discovery techniques.
These methods use conditional independence tests between variables in order to uncover the causal structure. To do so, they rely on two main assumptions: that the graph fulfills the global Markov property and the faithfulness assumption. Together they state that we observe conditional independence relations between two variables in the distribution if and only if these two variables are d-separated \cite{greiger_verma_pearl_bn_indep} in the graphical model.
For more details, we refer to \cref{appendix-causal-pre}. 

Causal inference is commonly approached with do-calculus \citep{pearl-do-calculus}. We denote the intervention distribution ${\mathbb P(Y|do(X=x))}$ to be the distribution of the outcome $Y$ conditioned on an intervention $do(X=x)$ which forces the treatment variable $X$ to take on the value $x$.
Note that this is in general not necessarily equal to ${\mathbb P(Y|X=x)}$.\footnote{For instance, there is a causal effect of altitude on temperature but not vice versa. Hence, intervening on temperature will not cause a shift in the distribution of altitude, but conditioning will.}
When they are not equal, we say that there is \textit{confounding}. Confounding occurs when there is a third variable $Z$, which causes both the treatment $X$ and the outcome $Y$. 

We say that there is a causal effect of $X$ on $Y$ if there exist $x$ and $x'$ such that
\begin{equation}
    \mathbb P(Y|do(X=x))\neq \mathbb P(Y|do(X=x')).
\end{equation}
One way to quantify the causal effect is with the \textit{average causal effect (ACE)}:
\begin{equation}
    \mathbb E[Y|do(X=x)]-
    \mathbb E[Y|do(X=x')].
\end{equation}

To estimate the causal effect using observational data, we need to rewrite the intervention distribution using only conditional distributions. Assuming a causal DAG, this can be done with the \textit{truncated factorization formula} \cite{pearl2009causal},
\begin{align}
\begin{split}
    & \mathbb P(X_{V}|do(X_{W}=x_{W}))=\\
    &=\prod_{i\in V\backslash W}\mathbb P(X_i|X_{pa(i)})\mathbbm 1_{\{X_{W}=x_{W}\}},
\end{split}
\end{align}
for $W\subset V$, with $X_{W}$ being the variables in $P$ corresponding to the nodes in $W$. 
\subsection{Causality for Change Dynamics
}
In this work, we 
estimate the direct causal effect of a word's type on its semantic change and frequency shift dynamics. 
In order to establish that such an effect exists, and to know which variables to control for, we turn to causal discovery algorithms. The variables in our causal graph additionally include frequency, polysemy and POS.

For learning the causal graph, we choose the constraint-based PC-stable algorithm \cite{colombo_maathuis_14}, an order-independent variant of the well-known PC algorithm \cite{spirtes2000causation}, discussed in \cref{appendix-causal-pre}. 
We are learning a mixed graphical model \cite{lauritzen_graphical_models, lee_hastie_graphical_models}, consisting of both continuous (e.g., frequency) and categorical (e.g., type) variables. For this reason we opt for constraint-based algorithms, allowing us to tailor the conditional independence tests according to the various data types. 

Having learned the causal graph (\cref{sec:result_causal_structure}), we proceed to estimate the ACE of word type on both semantic change and frequency shift using do-calculus (\cref{sec:causal_effect}).

\section{Slang and Nonslang Word Selection}
We select 100 slang words and 100 nonslang words for our study, presented in \cref{appendix-selected-words}. In the trade-off between 
statistical significance
and time spent on computation and data collection, we found that a set of 200 words was enough to get highly significant results. 
The slang words are randomly sampled from the Online Slang Dictionary,\footnote{\url{http://onlineslangdictionary.com/}} which provides well-maintained and curated slang word definitions as well as a list of 4,828 featured slang words as of June 2021.
We limit the scope of our study to only encompass single-word expressions, and in so doing we filter out 2,169 multi-word expressions.
To further clean the data, we also delete words with only one character and acronyms. 
Lastly, we limit the causal analysis to words that are exclusively either slang or nonslang, excluding ``hybrid'' words with both slang and nonslang meanings, such as ``kosher''
or ``tool.'' Including words of this type would have interfered with the causal analysis by creating a hardcoded dependency between word type and polysemy, as these words by definition are polysemous. 
We do however perform a separate analysis of the hybrid words in \cref{appendix-hybrid}. 

For the reference set of standard, nonslang, words we sample 100 words uniformly at random
from a list of all English words, supplied by the \texttt{wordfreq} library in Python \cite{wordfreq}.

\section{Data Collection
}
We curate a Twitter dataset from the years 2010 and 2020, which we select as our periods of reference, 
and collect the following variables:
\begin{itemize}[leftmargin=*]
  \setlength\itemsep{0.1em}
    \item \textbf{Word type:} Whether a word is slang or not
    \item \textbf{Word frequency:} The average number of tweets containing the word per day in 2010 and 2020
    (\cref{sec:freq})
    \item \textbf{Frequency Shift:} The relative difference in frequency the word has undergone between 2010 and 2020 (\cref{sec:freq_change})
    \item \textbf{Polysemy:} 
    The number of senses a word has (\cref{sec:polysemy})
    \item \textbf{Part of speech:} A binary variable for each POS tag (\cref{sec:pos})
    \item \textbf{Semantic change:} The semantic change score of the word from 2010 to 2020 (\cref{sec:semantic})
\end{itemize}

\subsection{Twitter Dataset}
\label{twitter-dataset}
As a social media platform, Twitter data is rich in both slang and nonslang words. The Twitter dataset we curated comprises 170,135 tweets from 2010 and 2020 that contain our selected words. 
Sampling tweets from two separate time periods 
allows us to examine the semantic change over a 10-year gap. For every slang and nonslang word, and each of the two time periods, we obtain 200-500 random tweets that contain the word and were posted 
during the corresponding year. 
We keep each tweet's text, tweet ID, and date it was posted.
As a post-processing step, we remove all 
URLs and hashtags from the tweets. To protect user privacy, we further replace all user name handles with the 
word ``user.''
On average, we have 346 tweets per slang word and 293 tweets per nonslang word. 

\subsection{Word Frequency}\label{sec:freq}

We approximate a word's frequency by the average number of times it is tweeted within 24 hours. This average is calculated in practice over 40 randomly sampled 24 hour time frames in a given year, in each of which we retrieve the number of tweets containing the word.
The frequencies are calculated separately for 2010 and 2020.
Due to the growing popularity of social media, the number of tweets has significantly increased over the decade. Therefore, we divide the counts from 2020 by a factor of $6.4$, which is the ratio between the average word counts in both years in our dataset. The frequencies from both years are then averaged to provide the \textit{frequency} variable for the causal analysis. 
\subsection{Frequency Shift}
\label{sec:freq_change}
\begin{figure}[t]
    \centering
\includegraphics[width=0.49\textwidth]{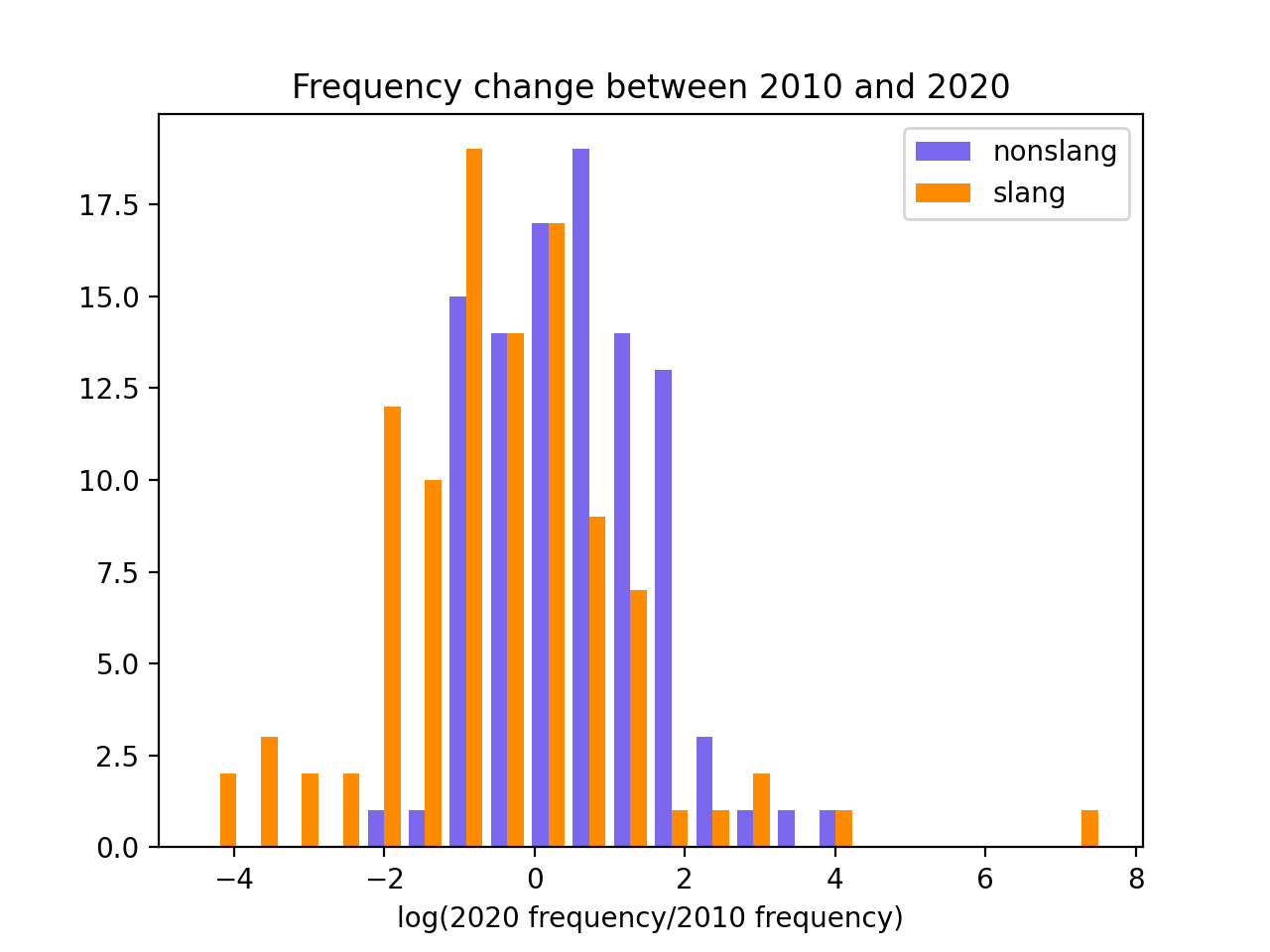}
\caption{Relative shift in frequency from 2010 to 2020, 
where a positive score corresponds to an increase in frequency. We see that slang words present both the highest increases and the highest decreases in frequency. Moreover, a large frequency decrease is observed exclusively in a set of slang words, indicating these words faded from usage during the decade.
}
\label{fig:logfreq}
\end{figure}

We are now interested in analyzing the dynamics of frequency shifts. To evaluate the relative change in frequency for a given word $w$ we take 
\begin{equation}
   \text{FreqShift}(w) = \log \frac{x_{2020}(w)}{x_{2010}(w)}
   \label{eq:logfreqchange}
\end{equation}
where, $x_k(w)$ is the frequency of word $w$ in year $k$. 
This has been shown to be the only metric for relative change that is symmetric, additive, and normed \cite{log_freq_change_measure}. Importantly, this measure symmetrically reflects both increases and decreases in relative frequency. 
The mean relative changes in frequency were 
 $-0.486 (\pm 1.644)$
for slang words and $0.533 (\pm 1.070)$ for nonslang words, where a positive score corresponds to an increase in frequency. As evident in \cref{fig:logfreq}, not only did more slang words exhibit a decrease in frequency than nonslang ones, the words that showed the highest frequency increase are also slang. 

We also examine the absolute value of 
\cref{eq:logfreqchange} to evaluate the degree of change, may it be a decrease or an increase. 
We find that, as expected, slang words have significantly higher changes in frequency than nonslang words ($p < 0.05$). See \cref{appendix-hybrid} for more details. 
\subsection{Polysemy}\label{sec:polysemy}
We define a word's polysemy score as the number of distinct senses it has\footnote{Note that this definition also encapsulates potential cases of homonymy. We choose not to make a distinction between polysemy and homonymy in this analysis.}. For nonslang words, we take the number of senses the word has in Merriam Webster
and for slang words we take the number of definitions on the Online Slang Dictionary. We use two separate resources as we find that no dictionary encapsulates both slang and nonslang words.
The mean polysemy scores are $(2.074 \pm 2.595)$ for slang words and $(3.079\pm2.780)$ for nonslang words with a significant difference in distribution $(p<0.05)$ according to a permutation test, implying that the latter are used with a larger variety of meanings. In addition, the slang senses of the hybrid words exhibit a distribution similar to those of the slang words (\cref{appendix-hybrid}).
More polysemous words tend to have a higher word frequency in our dataset -- the log transform of frequency and polysemy display a highly significant ($p<0.001$) linear correlation coefficient of $0.350$.

\subsection{Part of Speech}
\label{sec:pos}
For each word, we retrieve four binary variables, indicating whether a word can be used as noun, verb, adverb or adjective, which were the four major POS tags observed in our data. 
To calculate these variables we run the NLTK POS tagger \cite{loper_bird_nltk} on the tweets, and collect the distribution of POS tags for each word. 
Note that a word may have more than one POS tag, depending on the context in which it is used. Each of the binary variables is then set to be 1 if the word had the corresponding POS tag in at least 5\% of its tweets and 0 otherwise. 
\subsection{Semantic Change Score}\label{sec:semantic}
In this section we explain the details of how we obtain the semantic change scores. We start by fine-tuning a bi-directional language model on a slang-dense corpus (\cref{sec:fine_tuning}), after which we survey the literature and propose metrics (\cref{sec:experiment_method}) that we use to perform an extensive experimentation study to find the most suitable one (\cref{sec:sem_eval_res}). Finally, we apply this metric to our sets of slang and nonslang words on the Twitter data (\cref{sec:sem_change_twitter}). 
\subsubsection{Obtaining Contextualized Representations}
\label{sec:fine_tuning}
We familiarize the bi-directional language model with slang words and the contexts in which they are used by fine-tuning it on the masked language modeling task. For this purpose we use a web-scraped dataset from the Urban Dictionary, previously collected by \citet{wilson-etal-2020-urban}.
After preprocessing and subsampling, the details of which can be found in \cref{appendix-ud-preprocessing}, we are left with a training set of $200,000$ slang-dense text sequences.

As our bi-directional language model we select RoBERTa \cite{roberta}.
Beyond performance gains compared to the original BERT \cite{bert}, we select this model since it allows for more subword units. 
We reason, that this could be useful in the context of slang words since potentially some of the sub-units used in these words would not have been recognized by BERT. We choose the smaller 125M parameter base version for computational reasons. 
We train the model using the Adam optimizer \cite{kingma2017adam} with 
different learning rates $\gamma$. The lowest loss on the test set was found with $\gamma=10^{-6}$, which we proceed with for scoring semantic change.
For more details on training configurations, we refer to \cref{appendix-fine-tuning}.

\subsubsection{Quantifying Semantic Change}
\label{sec:experiment_method}

In order to select a change detection metric, we evaluate our model on the SemEval-2020 Task 1 on Unsupervised Lexical Semantic Change Detection \cite{schlechtweg2020semeval1}. This task provides the first standard evaluation framework for semantic change detection, using a large-scale labeled dataset for four different languages. We restrict ourselves to English and focus on subtask 2, which concerns ranking a set of $37$ target words according to their semantic change between two time periods. The ranking is evaluated using Spearman’s rank-order correlation coefficient $\rho$.\footnote{We note the caveat that our model is fine-tuned on Urban Dictionary text, while the older of the two English datasets of SemEval consists of text from 1810-1860. 
} Our space of configurations includes layer representations, dimensionality reduction techniques and semantic change metrics. 
\paragraph{Layer Representations:} 
Previous work \cite{ethayarajh2019contextual} has shown that embeddings retrieved from bi-directional language models are not isotropic, but are rather concentrated around a high-dimensional cone. Moreover, the level of isotropy may vary according to the layer from which the representations are retrieved \cite{ethayarajh2019contextual, cai2021isotropy}. This leads us to experiment with representations from different layers in our fine-tuned RoBERTa model, namely, taking only the first layer, only the last layer or summing all layers.
\paragraph{Dimensionality Reduction:} 
To the best of our knowledge, only one previous semantic change detection approach \cite{rother-etal-2020-cmce} has incorporated dimensionality reduction, more specifically UMAP \cite{mcinnes2018umap}. 
As the Euclidean distances in the UMAP-reduced space are very sensitive to hyperparameters and it does not retain an interpretable notion of absolute distances, it might be unsuitable for pure distance-based metrics like APD, and we therefore also experiment with PCA.

\paragraph{Metrics for Semantic Change:}
Given representations $\mathcal X_{t}=\{\bm {x}_{1,t},..., \bm {x}_{n_t,t}\}$ for a particular word in time period $t$, we define the average pairwise distance (APD) between two periods as
\begin{align}
\mathrm{APD}(\mathcal X_{t_1}, \mathcal X_{t_2})=\frac{1}{n_{t_1} n_{t_2}}\sum_{\substack{\bm {x}_{i,t_1}\in\mathcal X_{t_1} \\ \bm {x}_{j,t_2}\in\mathcal X_{t_2}}}d(\bm {x}_{i,t_1}, \bm {x}_{j,t_2})
~,
\end{align}
for some distance metric $d(\cdot,\cdot)$, where $n_{t_1}, n_{t_2}$ are the number of words in each time period. We experiment with Euclidean distance $d_2(\bm {x}_1, \bm {x}_2)$, cosine distance $d_{\mathrm{cos}}(\bm {x}_1, \bm {x}_2)$ and Manhattan distance $d_1(\bm {x}_1, \bm {x}_2)$. Furthermore, we propose a novel combined metric. Note that $d_2(\cdot, \cdot)\in[0,\infty]$ and $d_{\mathrm{cos}}(\cdot, \cdot)\in[0,2]$. Further note that 
\begin{align}||\bm {x}_1 - \bm {x}_2||^2_2
& \leq||\bm {x}_1||_2^2+ ||\bm {x}_2||_2^2
\end{align}
Normalizing both metrics for a support in $[0,1]$, we get a combined metric with the same unit support to be the following average:
\begin{align}
d_{2,\mathrm{cos}}(\bm {x}_1, \bm {x}_2) & =\frac{0.5\cdot d_2(\bm {x}_1, \bm {x}_2)}{\sqrt{||\bm {x}_1||^2 +||\bm {x}_2||^2}}\\
&+\frac{d_{\mathrm{cos}}(\bm {x}_1, \bm {x}_2)}{4}
\end{align}
We argue that this provides a more complete metric, capturing both absolute distance and the angle between vectors. 

In addition to the APD metrics, we experiment with distribution-based metrics (see \cref{appendix-distribution-metrics}).

\subsubsection{Evaluating the Semantic Change Scores
}
\label{sec:sem_eval_res}


We first compare the results for the three types of layer representations for different APD metrics, and note that summing all layer representations yields the best results. 
Consequentially, we proceed with the rest of the experiments using only these representations. For both PCA and UMAP, we experiment with projecting the representations down to $h\in\{2,5,10,20,50,100\}$ dimensions. These combinations are tested together with 
the APD metrics as presented in \cref{sec:experiment_method} as well as the distribution-based metrics described in \cref{appendix-semeval}. The latter do not however in general display significant correlations. 
\begin{table}[t]
    \centering
    \begin{tabular}{cccc}
    \toprule
        \textbf{Reduction} & $\bm{h}$ & \textbf{APD} & \textbf{Score} \\\midrule
        PCA & $100$ & $d_2$ and $d_{\mathrm{cos}}$ & $\bm{0.489}^{**}$ \\
        PCA & $100$ & $d_{\mathrm{cos}}$ & $0.464^{**}$\\
        PCA & $100$ & $d_2$ & $0.298$ \\
        None & $768$ & $d_2$ and $d_{\mathrm{cos}}$ & $0.345^*$ \\ 
    \bottomrule
    \end{tabular}
    \caption{Spearman's rank-order correlation coefficients between our semantic change scores and the ground truth across different dimensionality reduction techniques for APD (*: $p<0.05$, **: $p<0.01$).}
    \label{tab:APD-results}
\end{table}

We present a small subset of the scores resulting from the APD configurations in \cref{tab:APD-results}, highlighting our finding that both PCA dimensionality reduction and using a combined metric improve the performance. More results and comparisons to baselines are presented in \cref{appendix-sem-eval-results}. 
We observe that the proposed combined metric consistently outperforms both $d_2$ and $d_{cos}$ across values of $h$ for PCA. We also note that UMAP projections perform poorly with the APD metrics and that
projecting down to 50-100 dimensions seems to be optimal, which maintains 70-85\% of the variance as we illustrate in \cref{appendix-visualizations-pca}. In addition, both norm-based metrics $d_1$ and $d_2$ perform worse with dimensionality reduction. As our final metric, we choose the best performing configuration on SemEval, with PCA $h=100$ and the combined metric, as seen in \cref{tab:APD-results}.

\subsubsection{Semantic Change Scores for Slang and Nonslang Words on the Twitter Dataset}
\label{sec:sem_change_twitter}
\begin{figure}[t]
    \centering
    \includegraphics[width=0.49\textwidth]{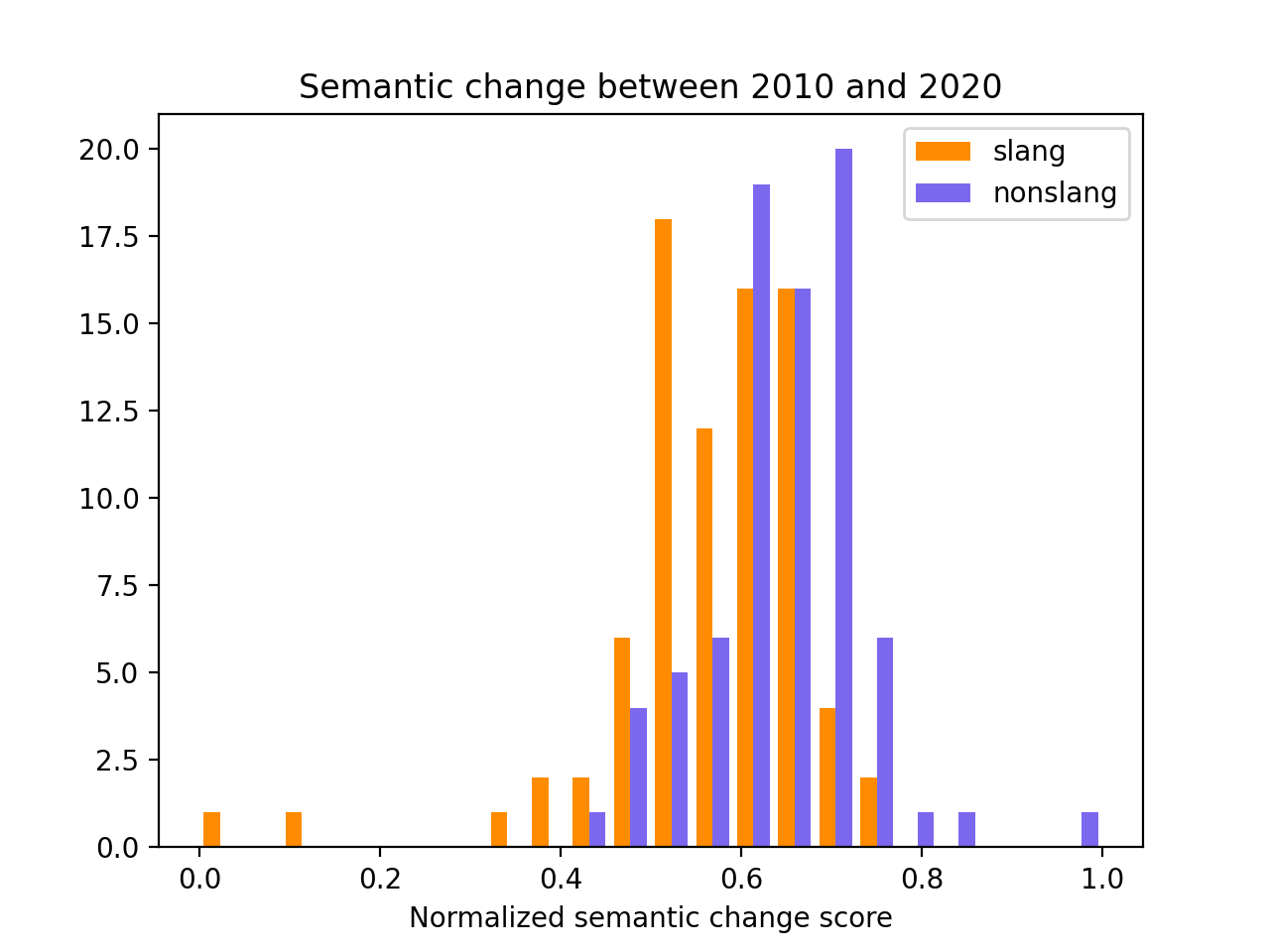}
    \caption{Semantic change scores between 2010 and 2020.
    We see that nonslang words typically underwent larger changes in meaning throughout the decade. 
    }
    \label{fig:APDscores}
\end{figure}

We obtain semantic change scores using the Twitter dataset described in \cref{twitter-dataset}. For the semantic change analysis, we exclude words that have less than 150 tweets in each time period within the dataset, which leaves us with 80 slang and 81 nonslang words.
We also normalize the scores according to the sample. The resulting semantic change scores are shown in \cref{fig:APDscores}. The mean semantic change scores are $0.564(\pm 0.114)$ for slang words and $0.648(\pm0.084)$
for nonslang words. The difference in semantic change score distributions is significant ($p < 0.001$) via a permutation test. The word with the highest semantic change score of $1$ is ``anticlockwise,'' and the word with the lowest score of $0$ is ``whadja.''

\section{Causal Analysis}
\subsection{Preparation for Causal Discovery}
PC-stable is constraint-based and thus makes use of conditional independence tests. In the case of continuous Gaussian variables, we can perform partial correlation tests to assess conditional independence, since zero partial correlation in this case is equivalent to conditional independence \cite{partial_correlation_conditional_independence}.
As word frequency has been suggested to follow a lognormal distribution \cite{wordfreq_distribution_statistical_models}, we take the log transform of it. The continuous variables \textit{semantic change}, \textit{frequency change} and \textit{log-frequency} are then all assumed to be approximated well by a Gaussian distribution, which is confirmed by diagnostic density and Q-Q plots (displayed in \cref{appendix-diagnostic}). 
\begin{figure*}[t]
    \centering
    \includegraphics[width=0.7\textwidth]{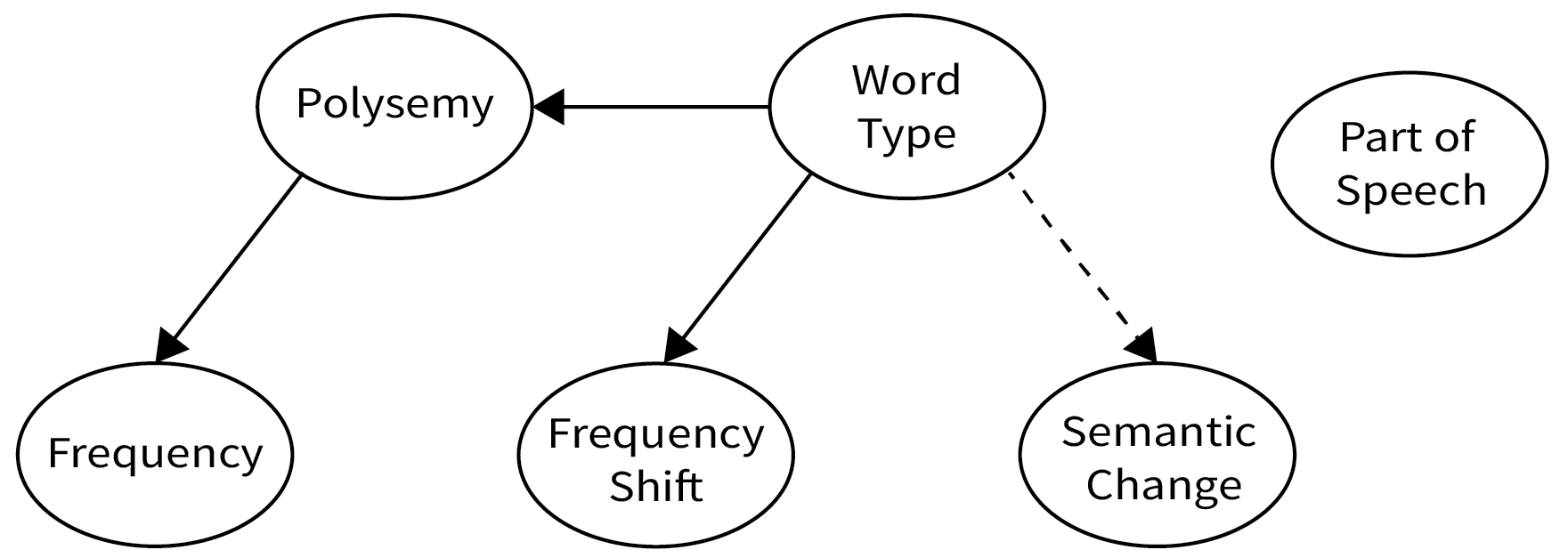}
    \caption{DAG representing the causal relationships in our dataset. We see that word type directly influences frequency shift, semantic change and polysemy, and polysemy in turn influences frequency.
    }
    \label{fig:causal-graph}
\end{figure*}

We categorize the discrete polysemy variable, experimenting with nine different plausible categorizations for the sake of robustness of the results.
Word type and POS are categorical in nature. For the categorical variables and for mixes of categorical and continuous variables, we perform chi-squared mutual information based tests \cite{edwards}, since the approximate null distribution of the mutual information is chi-squared \cite{brillinger_data_analysis}. For all conditional independence tests we experiment with significance levels $\alpha\in\{0.01, 0.03, 0.05\}$.

\subsection{Resulting Causal Structure}
\label{sec:result_causal_structure}

In \cref{fig:causal-graph} we see the result from the above approach, with dashed lines representing edges that were apparent in most but not all of the configurations.
See \cref{appendix-causal-structure} for a sensitivity analysis.
\newline\indent
We first observe that word type has a direct causal effect on both the semantic change score and the frequency shift, without any confounding from the other variables. We also note a direct influence of word polysemy on frequency.
\newline\indent
Moreover, none of the four POS categories, which are all gathered in one node in \cref{fig:causal-graph}, have a causal link to any of the other variables.
We additionally observe a dependency between word type and polysemy. 
This edge could not be oriented by the PC-stable algorithm, however we manually orient it as outgoing from type and ingoing to polysemy, since an intervention on type should have a causal effect on the number of word senses and not vice versa. It is also interesting to note that polysemy does not seem to have a causal effect on semantic change. Its association with semantic change ($p<0.05$, rejecting the null hypothesis of independence between polysemy and semantic change) is instead confounded by word type. 

\subsection{Causal Effects
}
\label{sec:causal_effect}
In our case of no confounders, evaluating the ACE of word type on semantic change 
is straight-forward, as it reduces to the difference between the conditional expectations:
\begin{align}
\begin{split}
    \mathbb E[S|do(T=\text{nonslang})]-
    \mathbb E[S|do(T=\text{slang})]=& \\
    =\mathbb E[S|T=\text{nonslang}]-
    \mathbb E[S|T=\text{slang}]& 
\end{split}
\end{align}
See \cref{appendix-causal-inference} for a derivation. The case of frequency shift is analogous. 

We estimate the expectations by the sample means on the normalized values and get an average causal effect of $0.084$, which is a highly significant value ($p<0.001$) based on a t-test. 
For the observed changes in relative frequency, calculated according to \cref{eq:logfreqchange}, we get an average causal effect of $1.017$ ($p<0.001$ via a t-test).


\section{Discussion}
We analyze the dynamics of frequency shift and semantic change in slang words, and compare them to those of nonslang words. Our analysis shows that {\bf slang words change slower in semantic meaning, but adhere to more rapid frequency fluctuations, and are more likely to greatly decrease in frequency}. Our study is the first computational approach to confirm this property in slang words \cite{ingroup_language_college_students, peoples_poetry}.

To ensure that this is the result of a causal effect, and not mediated through another variable or subject to confounders, we model the data with a causal DAG, by also considering the potential interacting variables polysemy, frequency and POS. We discover that there is no influence of confounders, nor are there mediators between a word's type and its semantic change or its frequency shift, which {\bf confirms a direct causal effect}. This means that if we could intervene on a word's type, i.e., by setting it to be slang instead of nonslang or vice versa, we would expect its change dynamics to differ. 

Our results are consistent with those of \citet{dubossarsky-etal-2017-outta}, which found that neither the law relating semantic change to frequency, polysemy \cite{hamilton-etal-2016-diachronic} nor  prototypicality \cite{Dubossarsky2015ABU} were found to be as strong as previously thought after a case-control study using a scenario without semantic change. Indeed, there is no directed path from polysemy or frequency to semantic change in our causal graph, but they are both influenced by word type. We leave for future research to explore whether other word
categorizations, e.g., related to specific domains, languages or phonetic aspects, sustain this result. 
\linebreak \indent
In addition, our analysis does not support the claim that POS could underlie semantic change
\cite{Dubossarsky2016VerbsCM}. We note however that as our vocabulary contains $50\%$ slang words, the results need not be consistent with results obtained with a word sample drawn from standard language. 

Moreover, in the causal structure we discover that {\bf word polysemy has a direct effect on word frequency}, which is in line with previous linguistic studies showing that a word's frequency grows in an S-shaped curve when it acquires new meanings \cite{kroch_1989_scurve, Feltgen_2017}, as well as a known positive correlation between polysemy and frequency \cite{evolution_polysemous_words, polysemy_brevity_vs_frequency}.
We emphasize that this relationship is not merely an artifact of contextualized word representations being affected by frequency \cite{zhou_frequency_distortion_contextualized}, since our polysemy score does not rely on word representations as in \citet{hamilton-etal-2016-diachronic}. Our approach is however not without drawbacks -- the polysemy variable is collected from dictionaries, which may be subjective in their assignments of word senses.

Our study, along with previous work on the dynamics of semantic change, is limited by mainly considering distributional factors. Linguists have suggested that sociocultural, psychological and political factors may drive word change dynamics \cite{blank-1999, bochkarev_lexical_change}, and slang words are not an exception. 
Although challenging to measure, the influence of such factors on slang compared to nonslang words would be interesting to examine in future work. 

In conclusion, we believe that a causal analysis as we have presented here provides a useful tool to understand the underlying mechanisms of language. Complementing the recent emergence of research combining causal inference and NLP \cite{feder2021causal}, we have shown that tools from causality can also be beneficial for gaining new insights in diachronic linguistics. 

\section{Conclusion}
In this work, we have analyzed the diachronic mechanisms of slang language with a causal methodology. This allowed us to establish that a word’s type has a direct effect on its semantic change and frequency shift, without mediating effects from other distributional factors.

\section*{Acknowledgments}
We would like to thank Steven R. Wilson for providing us with the Urban Dictionary data and Walter Rader for providing us with a curated set of slang words from the Online Slang Dictionary. For the Twitter data, we are thankful to have been able to get access to Twitter's Academic Research Track. Finally, we gratefully acknowledge feedback and helpful comments from Mario Giulianelli, Yifan Hou, Bernhard Schölkopf and three anonymous reviewers.

This material is based in part upon works supported by the John Templeton Foundation (grant \#61156); by a Responsible AI grant by the Haslerstiftung; by an ETH Grant
(ETH-19 21-1); by the German Federal Ministry of Education and Research (BMBF): Tübingen AI Center, FKZ: 01IS18039B; and by the Machine Learning Cluster of Excellence, EXC number 2064/1 – Project number 390727645.

\section*{Ethical Considerations}
Our dataset is composed solely of English text. This means that our analysis applies uniquely to the English language, and results may differ in other languages. Moreover, for the purpose of this study, we curated a dataset of $170,135$ tweets. We emphasize that in order to protect the anonymity of users, we remove all author IDs from the data, and replace all usernames with the general token ``user.'' In the Urban Dictionary dataset we received from \citet{wilson-etal-2020-urban}, we similarly remove the author IDs and only consider the entry text. 

\bibliography{custom}
\bibliographystyle{acl_natbib}

\appendix
\newpage
\section{Appendix -- Fine-tuning with Urban Dictionary data}
\subsection{Preprocessing}
\label{appendix-ud-preprocessing}
\begin{figure}[t]
    \centering
    \includegraphics[width=0.45\textwidth]{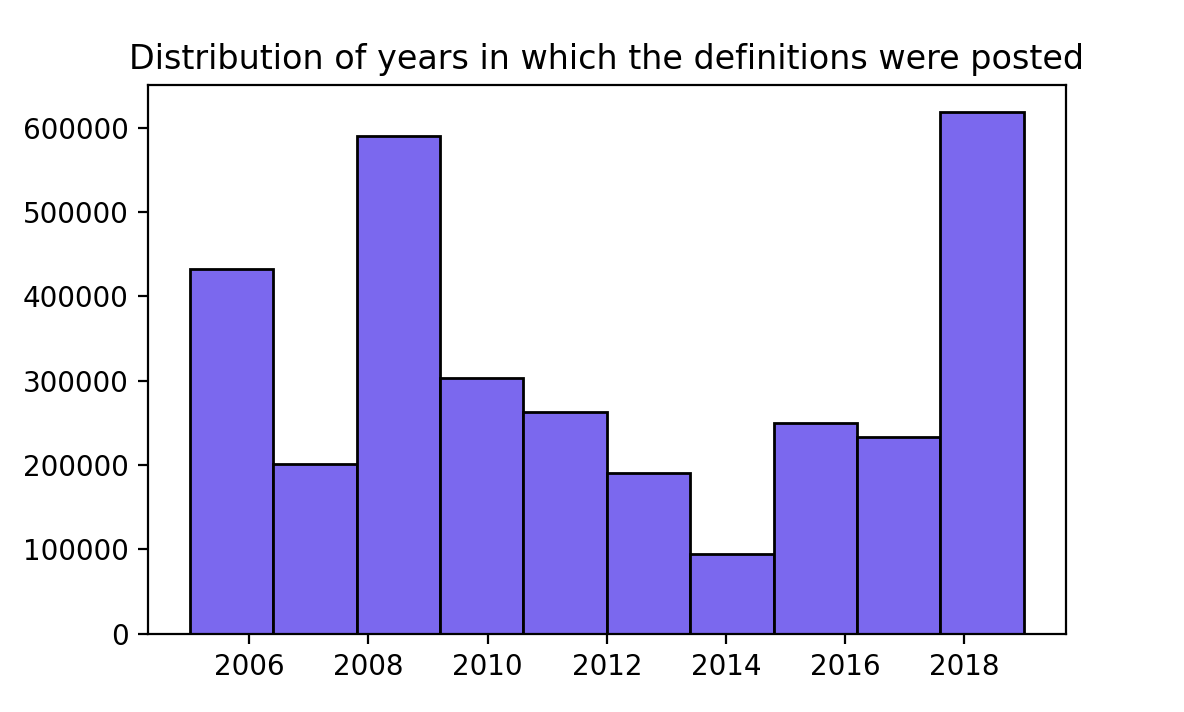}  
    \caption{Frequency counts over years in Urban Dictionary data}
    \label{fig:distrib_UD_year}
\end{figure}

The full Urban Dictionary data contains $3,534,966$ word definitions. In the dataset provided by \citet{wilson-etal-2020-urban}, each entry contains a definition, examples in which the word occurs, number of upvotes \& downvotes from website visitors, username of the submitter and a timestamp. As the data is crowd-sourced, many of these entries are noisy and of low quality. We therefore filter the lower quality definitions out 
before fine-tuning RoBERTa. 
After performing data exploration, we came up with two criteria that we found the most indicative of a definition's quality: the number of upvotes it got, and its upvote/downvote ratio. The distribution of upvotes, downvotes and the upvote/downvote ratios in the dataset can be seen in \cref{fig:distrib_UD} below. We also note that the number of submissions to Urban Dictionary is relatively well-spread, see \cref{fig:distrib_UD_year}. This implies that we do not have a strong bias towards more recently popularized slang terms in the dataset, and that we do have representation of the entire time span of interest; $2010-2020$.

\begin{figure}[t]
    \centering
    \includegraphics[width=0.4\textwidth]{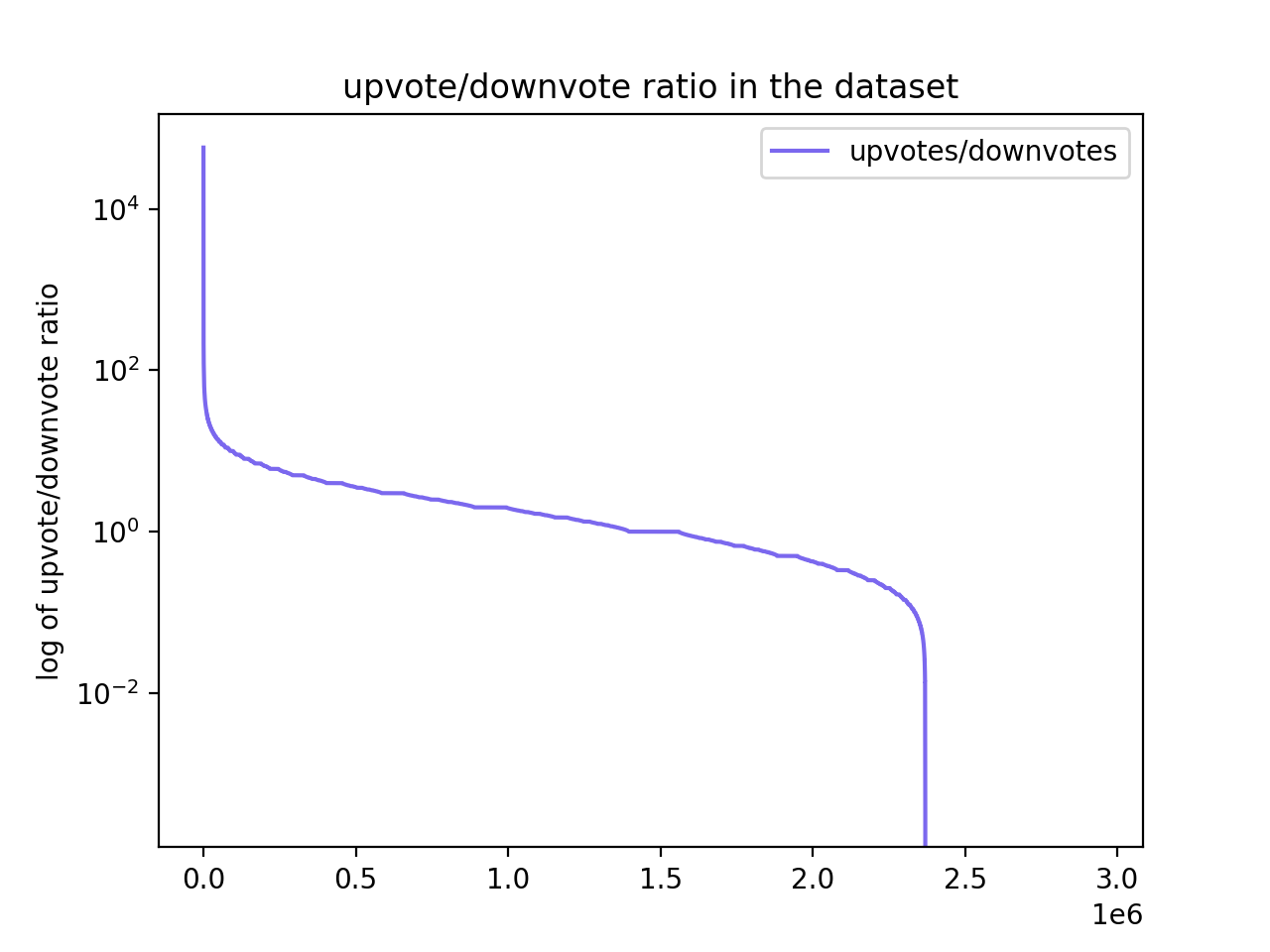}
    \includegraphics[width=0.4\textwidth]{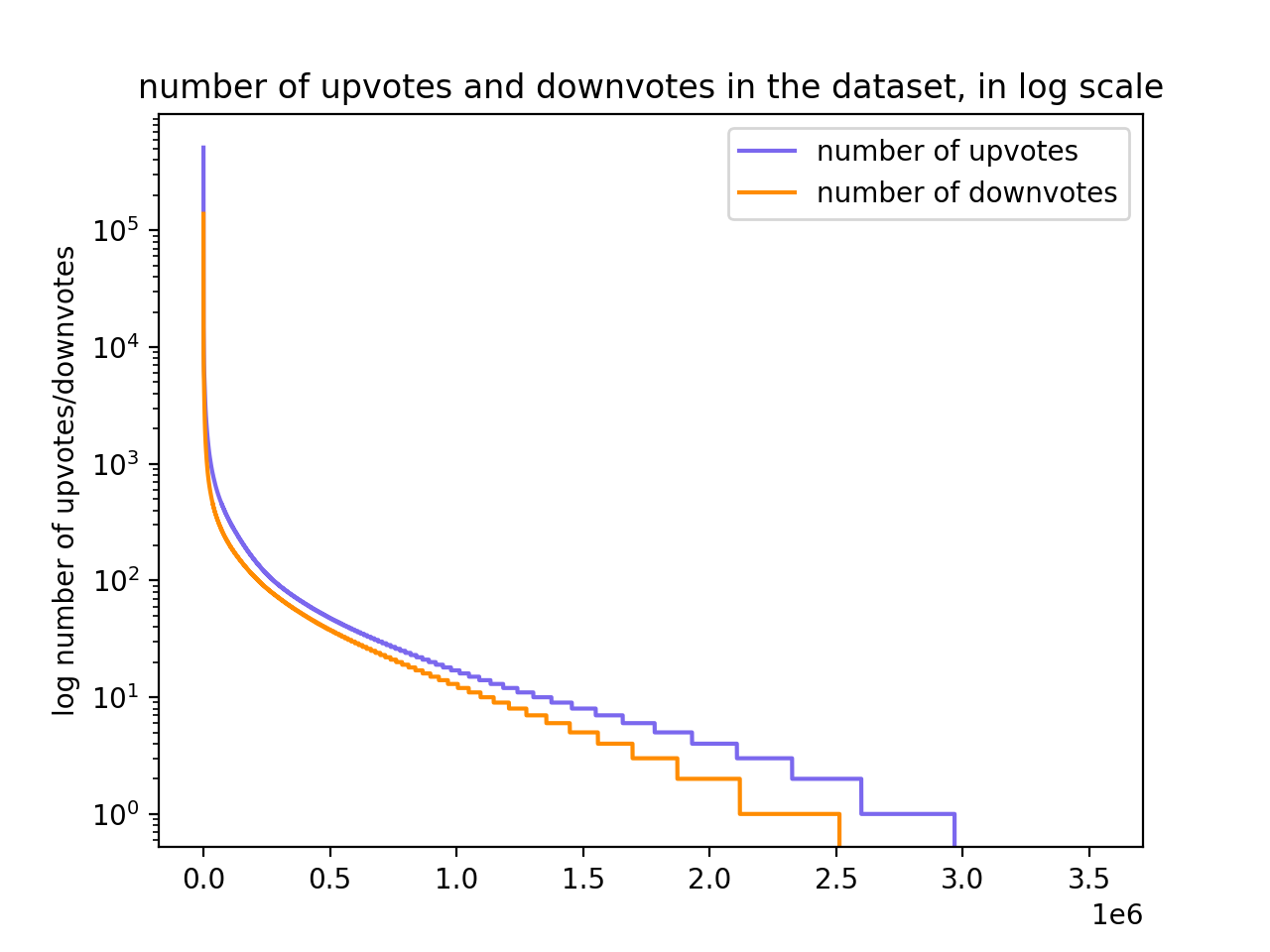}
    \caption{The distributions of (a) upvote/downvote ratio, (b) number of upvotes and number of downvotes among definitions in the dataset in log-scale.}
    \label{fig:distrib_UD}
\end{figure}

We keep the entries having more than $20$ upvotes and an upvote/downvote ratio of at least $2$. This leaves us with $488,010$ Urban Dictionary entries, out of which we randomly sample $100,000$ to reduce the computation time in the fine-tuning process. We use both the definitions and the word usage examples for fine-tuning, producing a final dataset of $200,000$ sequences.

\subsection{Training}
\label{appendix-fine-tuning}

We randomly split the data into $80\%$ train and $20\%$ test, before training for $10$ epochs with an early stopping with patience $3$. The batch size was set to $1$ in the interest of memory constraints.
Following the setup from the pre-training stage as explained in \citet{roberta}, we use the Adam optimizer \cite{kingma2017adam} with $\epsilon=10^{-6},\beta_1=0.9$ \& $\beta_2=0.98$ and a linear learning rate decay. For the learning rate, we argue that since the initialized parameters should provide a solution which is already close to the optimum when evaluating on our dataset (our fine-tuning being the very same masked language modeling task as RoBERTa has already been trained on), the learning rate should be smaller. Thus, instead of picking the learning rate $\gamma=6\cdot10^{-4}$ as was done by \citet{roberta}, we experiment with $\gamma\in\{10^{-4},10^{-5},10^{-6},10^{-7}\}$. Training was done using an NVIDIA GeForce GTX 1080 8GB GPU and took around 1 to 1.5 days per model. 

\newpage
\section{Appendix -- Experiments on SemEval-2020}
\label{appendix-semeval}
\subsection{Distribution-based Metrics}
\label{appendix-distribution-metrics}
\paragraph{Method:}In addition to the distance-based APD metrics, we experiment with two distribution-based ones, namely entropy difference (ED) \& Jensen-Shannon Divergence (JSD) \citep{giulianelli-etal-2020-analysing}. 

We assume a categorical distribution over a set of $K_w$ word senses for word $w$ and time period $t$. The word sense $s_i^w$ of an occurrence $i$ is then given by:
\[s_i^{wt}\sim Cat(\alpha^{wt}_1, ..., \alpha^{wt}_{K_w})=:P^{wt}\]
Given two time periods of word sense distributions, we define the ED metric as
\[|H(s^{wt_2})-H(s^{wt_1})|\]
with entropy $H(\cdot)$. The JSD is given as:
\[\frac{1}{2}KL(P^{wt_1}||M)+\frac{1}{2}KL(P^{wt_2}||M)\]
with $M=\frac{P^{wt_1}+P^{wt_2}}{2}$ and $KL(\cdot||\cdot)$ being the KL-divergence. 

We obtain the word sense distributions via a clustering of the representations from both time periods. We experiment with K-Means and Gaussian Mixture Models (GMMs), the latter proposed due to its ability to find more general cluster shapes. We also experiment briefly with Affinity Propagation, which has been used in previous semantic change detection work \cite{capturing_evolution_more_clusters, semeval_kutuzov, montariol-etal-2021-scalable}. However, we find it to be ill-suited for our purposes since it results in an excessive amount of clusters in comparison to how a human would classify word senses. 

For both K-means and GMM, we experiment with selecting the optimal $K_w\in[1,10]$ through two different procedures. The first one is a slight extension of the method from \citet{giulianelli-etal-2020-analysing} -- we select the $K_w$ which optimizes the silhouette score \cite{silhouette} for a set of different initializations. Their approach does not consider the single cluster case however
, so we extend it by setting $K_w=1$ when the best silhouette score is below a threshold of $0.1$. For $K$-Means, we further experiment with an automatic elbow method\footnote{See https://kneed.readthedocs.io/en/stable/index.html} for the sum of squared distances to the cluster centroids, which decreases monotonically with the number of clusters. We again select the cluster assignments with the largest silhouette score for multiple random initalizations. For GMM, we further experiment with taking the model which corresponds to the best Bayesian Information Criterion \cite{BIC}. 

\begin{figure}[t]
    \centering
    \includegraphics[width=0.45\textwidth]{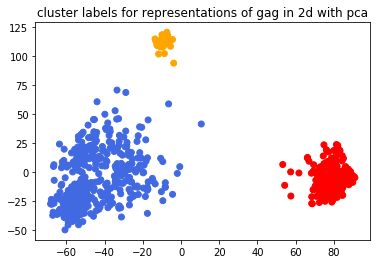}
    \caption{Clusters found with GMM from 2-dimensional PCA representations of the word \textbf{gag}.}
    \label{fig:gag}
\end{figure}

\paragraph{Clustering examples:}
\label{appendix-visualizations-clustering}
In \cref{fig:gag} we see three clusters found for ``gag.'' They do not seem to correspond to word senses however: An example from the first cluster is ``user i need a pic of you begging if i ' m boiling these because boiled eggs make me gag . :d,'' an example from the second cluster is ``lmao rt user user user so i tried that tuna with cheese and my gag reflexes were in full affect !'' and an example from the third cluster is ``gag me with a spoon'' -- all seemingly referring to the sensation of being about to vomit.

\begin{figure}[t]
    \centering
    \includegraphics[width=0.47\textwidth]{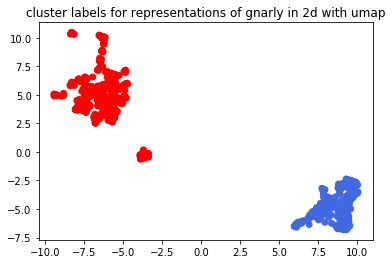}
    \caption{Clusters found with GMM from 2-dimensional UMAP representations of the word \textbf{gnarly}.}
    \label{fig:gnarly}
\end{figure}

We show another example in \cref{fig:gnarly} of the word ``gnarly,'' this time reduced to $2$ dimensions using UMAP. Gnarly has three meanings according to the Online Slang Dictionary: It can either mean very good / excellent / cool, gross / disgusting or painful / dangerous. These three word senses are not separated by UMAP and GMM, for instance both ``its a good thing one of my roomies is a dude , who else would kill gnarly spiders in my room when i start to hyperventilate'' and ``rt user bro my wreck on the scooter was so gnarly like it was fun i love shit like that . i wish i could’ve been on jackass'' are put in the first cluster.

\subsection{Variance Explained by PCA components}
\label{appendix-visualizations-pca}

\begin{figure}[t]
    \centering
    \includegraphics[width=0.48\textwidth]{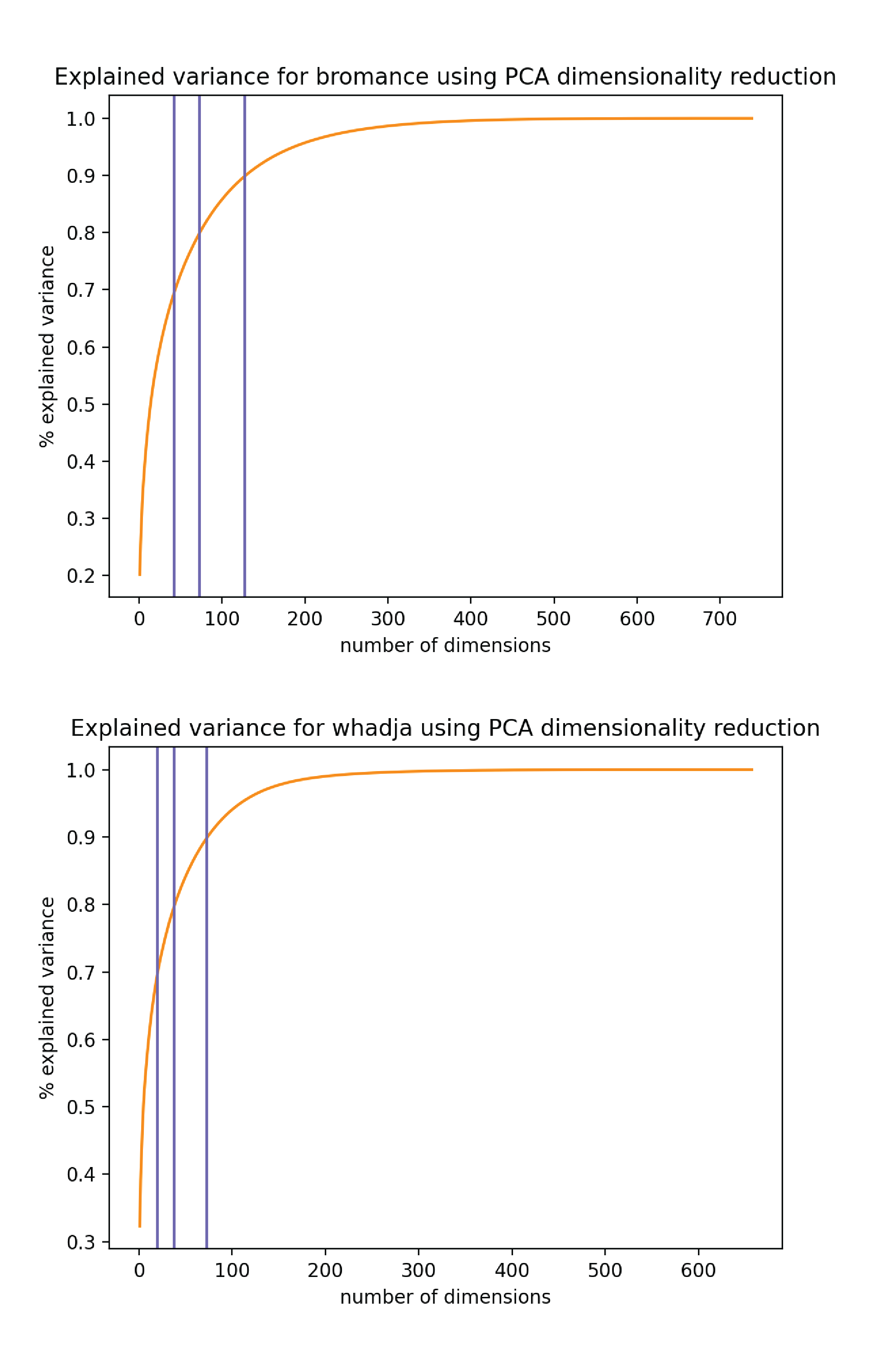}
    \caption{Explained variance by number of components used in PCA for the slang words \textit{bromance} and \textit{whadja}}
    \label{fig:pcavar}
\end{figure}

Consider \cref{fig:pcavar}
for example plots of how much variance is preserved with PCA on the contextualized representations.

\subsection{Results}
\label{appendix-sem-eval-results}
We further present more results of the experimentation on the SemEval-2020 Task 1 Subtask 2. All tables show the Spearman's rank-order correlation between the change metrics and the ground truths. 

In \cref{tab:baselines} we compare our best performing setup to the three best performing previous approaches on SemEval-2020 Task 1 Subtask 2. We see that only \citet{semeval_kutuzov} display a higher score, which might be partially explained by the fact that they fine-tune their model on the SemEval test corpora. We do not do this since our main goal is not to beat state-of-the-art on the shared task, but rather to find a good enough model to detect semantic change in slang.

The results comparing the layer representations can be observed in \cref{tab:layer-reps-results}. As a side observation we also note that the less isotropic first layer representations seem to perform better than the more isotropic last layer representations.

\begin{table}[t]
    \centering
    \begin{tabular}{cc}
    \toprule
        \textbf{Baseline} & \textbf{Score} \\\hline
        Combined APD PCA100 & $0.489$ \\
        \citet{semeval_kutuzov} & $0.605$ \\
        \citet{kaiser-et-al-semeval} & $0.461$ \\
        \citet{rother-etal-2020-cmce} & $0.440$ \\
    \bottomrule
    \end{tabular}
    \caption{Comparison to the three highest performing previous works on the SemEval-2020 Task 1 subtask 2 for the English dataset.}
    \label{tab:baselines}
\end{table}

\begin{table}[t]
    \centering
    \begin{tabular}{ccc}
    \toprule
         & \textbf{$d_2$ APD} & \textbf{$d_{\mathrm{cos}}$ APD} \\\midrule
        First layer &  $0.22$ & $0.234$ \\
        Last layer & $0.07$ & $0.2$ \\
        Sum of all layers & $\bm{0.336}^*$ & $\bm{0.332}^*$\\
    \bottomrule
    \end{tabular}
    \caption{Spearman's rank-order correlation coefficients between our semantic change scores and the ground truth across different layer representations ($p<0.05$).}
    \label{tab:layer-reps-results}
\end{table}
In \cref{tab:layer-change-scores} we present a comparison across different layer representations for both APD-based and distribution-based metrics. 
We observe that none of the distribution-based metrics give significant
results, even when used with dimensionality reduction techniques. While a few of them do have a slight positive correlation, we omit this approach altogether. The APD results on the other hand show a high correlation for many of the configurations, providing an indication of the APD's robustness in detecting semantic change. We show a selection of these in \cref{tab:APD-more-results}.
\begin{table}[t]
    \centering
    \begin{tabular}{ccccc}
    \toprule
        \textbf{Reps} & \textbf{Cluster} & \textbf{Metric} & \textbf{Score} & $\bm{p}$ \\\hline
        First & - & APD $d_2$ & $0.22$ & $0.19$\\
        First & - & APD $d_{\mathrm{cos}}$ & $0.23$ & $0.16$\\
        First & K-Means & ED & $-0.08$ & $0.64$ \\
        First & K-Means & JSD & $0.06$ & $0.73$ \\
        First & GMM & ED & $0.05$ & $0.76$ \\
        First & GMM & JSD & $0.07$ & $0.67$ \\
        \hline
        Last & - & APD $d_2$ & $0.01$ & $0.97$\\
        Last & - & APD $d_{\mathrm{cos}}$ & $0.20$ & $0.24$\\
        Last & K-Means & ED & $0.00$ & $0.96$ \\
        Last & K-Means & JSD & $0.20$ & $0.23$ \\
        Last & GMM & ED & $-0.07$ & $0.70$ \\
        Last & GMM & JSD & $-0.10$ & $0.57$ \\
        \hline
        All & - & APD $d_2$ & $0.34$ & $0.04$\\
        All & - & APD $d_{\mathrm{cos}}$ & $0.33$ & $0.05$\\
        All & K-Means & ED & $0.03$ & $0.85$ \\
        All & K-Means & JSD & $0.09$ & $0.60$ \\
        All & GMM & ED & $-0.13$ & $0.43$ \\
        All & GMM & JSD & $0.00$ & $0.99$ \\
    \bottomrule
    \end{tabular}
    \caption{Comparison across different layer representations with APDs and distribution metrics, with $K_w$ selected through silhouette scores.}
    \label{tab:layer-change-scores}
\end{table}

\begin{table}[t]
    \centering
    \begin{tabular}{ccc}
    \toprule
         \textbf{APD} & \textbf{Score} & $\bm p$ \\\hline
        $d_2$ & $0.336$ & $0.042$\\
        $d_{\mathrm{cos}}$ & $0.332$ & $0.045$\\
        $d_1$ & $0.409$ & $0.012$\\
        $d_2$ and $d_{\mathrm{cos}}$ & $0.345$ & $0.037$\\
        $d_2,d_{\mathrm{cos}}$ and $d_1$ & $0.398$ & $0.015$\\
    \bottomrule
    \end{tabular}
    \caption{Comparison across APD metrics for original representations. Representations are sums across all layers.}
\end{table}

\begin{table}[t]
    \centering
    \begin{tabular}{cccc}
    \toprule
        \textbf{Dim} & \textbf{APD} & \textbf{Score} & $\bm p$ \\\hline
        PCA2 & $d_2$ & $-0.153$ & $0.367$\\
        UMAP2 & $d_{\mathrm{cos}}$ & $-0.136$ & $0.424$\\
        PCA5 & $d_{\mathrm{cos}}$ & $0.209$ & $0.215$\\
        PCA5 & $d_2$ and $d_{\mathrm{cos}}$ & $0.268$ & $0.109$\\
        UMAP5 & $d_2,d_{\mathrm{cos}}$ and $d_1$ & $-0.146$ & $0.39$\\
        PCA20 & $d_2$ and $d_{\mathrm{cos}}$ & $0.42$ & $0.010$\\
        PCA50 & $d_2$ & $0.26$ & $0.121$ \\
        PCA50 & $d_{\mathrm{cos}}$ & $0.394$ & $0.016$ \\
        PCA50 & $d_2$ and $d_{\mathrm{cos}}$ & $0.478$ & $0.003$ \\
        PCA50 & $d_2,d_{\mathrm{cos}}$ and $d_1$ & $0.344$ & $0.037$\\
        UMAP50 & $d_2$ & $-0.158$ & $0.35$\\
        PCA100 & $d_1$ & $0.297$ & $0.074$\\
        PCA100 & $d_2$ and $d_{\mathrm{cos}}$ & $0.489$ & $0.002$\\
        UMAP100 & $d_{\mathrm{cos}}$ & $-0.133$ & $0.433$\\
    \bottomrule
    \end{tabular}
    \caption{Comparison across different dimensions with PCA and UMAP for APD metrics. Representations are sums across all layers.}
    \label{tab:APD-more-results}
\end{table}

\newpage

\section{Appendix -- Hybrid Words
}
\label{appendix-hybrid}
We define hybrid words as words that have both a slang and nonslang meaning, i.e., occurring in both Online Slang Dictionary (OSD) and Merriam Webster (MW). In this section, we compare the polysemy, semantic change, frequency shift as well as the absolute frequency change patterns of hybrid words to slang and nonslangs.

Polysemy is collected for hybrid words from OSD and MW separately. Since the MW dictionary may also contain slang meanings, we filter out definitions labeled as \textit{slang}, \textit{informal} or \textit{vulgar} from these scores. The mean polysemy scores of the slang words are $(2.074\pm2.568)$ and the mean OSD polysemy scores of the hybrid words are $(2.580\pm 2.178)$, with a non-significant difference ($p>0.05$) in distribution according to a permutation test. This tells us that we are not skewing the polysemy score distribution of the slang words by excluding hybrid words. 

As for the nonslang meanings of the hybrid words, we get a mean polysemy score of $(6.880\pm6.080)$ which is significantly different $(p<0.001)$ from those of the nonslang words $(3.079\pm2.780)$. This is an interesting observation, implying that had we included nonslang words with hybrid meaning in our nonslang words sample, the difference in polysemy between slang and nonslang words would have been larger. Some example words from this category with high MW polysemy scores include ``split,'' ``down'' and ``walk.''

\begin{figure}[t]
    \centering
    \includegraphics[width=0.48\textwidth]{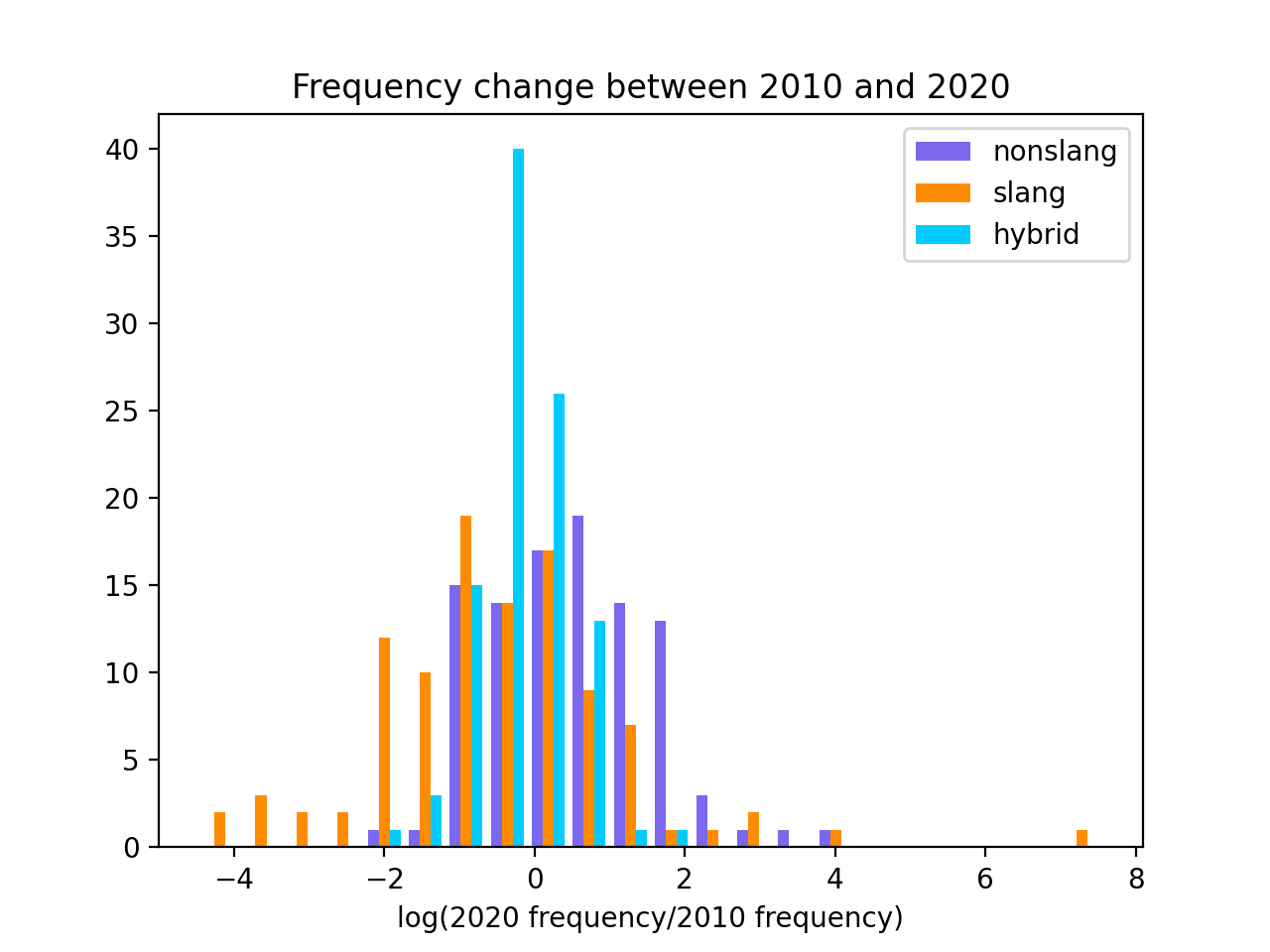}
    \caption{Relative difference in frequency between 2020 and 2010, for slang, nonslang and hybrid words, where a positive score corresponds to an increase in frequency.}
    \label{fig:hybrid-freq-change}
\end{figure}

For the relative frequency changes, we present the results as histograms in \cref{fig:hybrid-freq-change}. The frequency changes in hybrid words seem to fall between those of the slang words and the nonslang words. We observe a mean and standard deviation of $-0.154$ and $0.608$ respectively. 

\begin{figure}[t]
    \centering
    \includegraphics[width=0.48\textwidth]{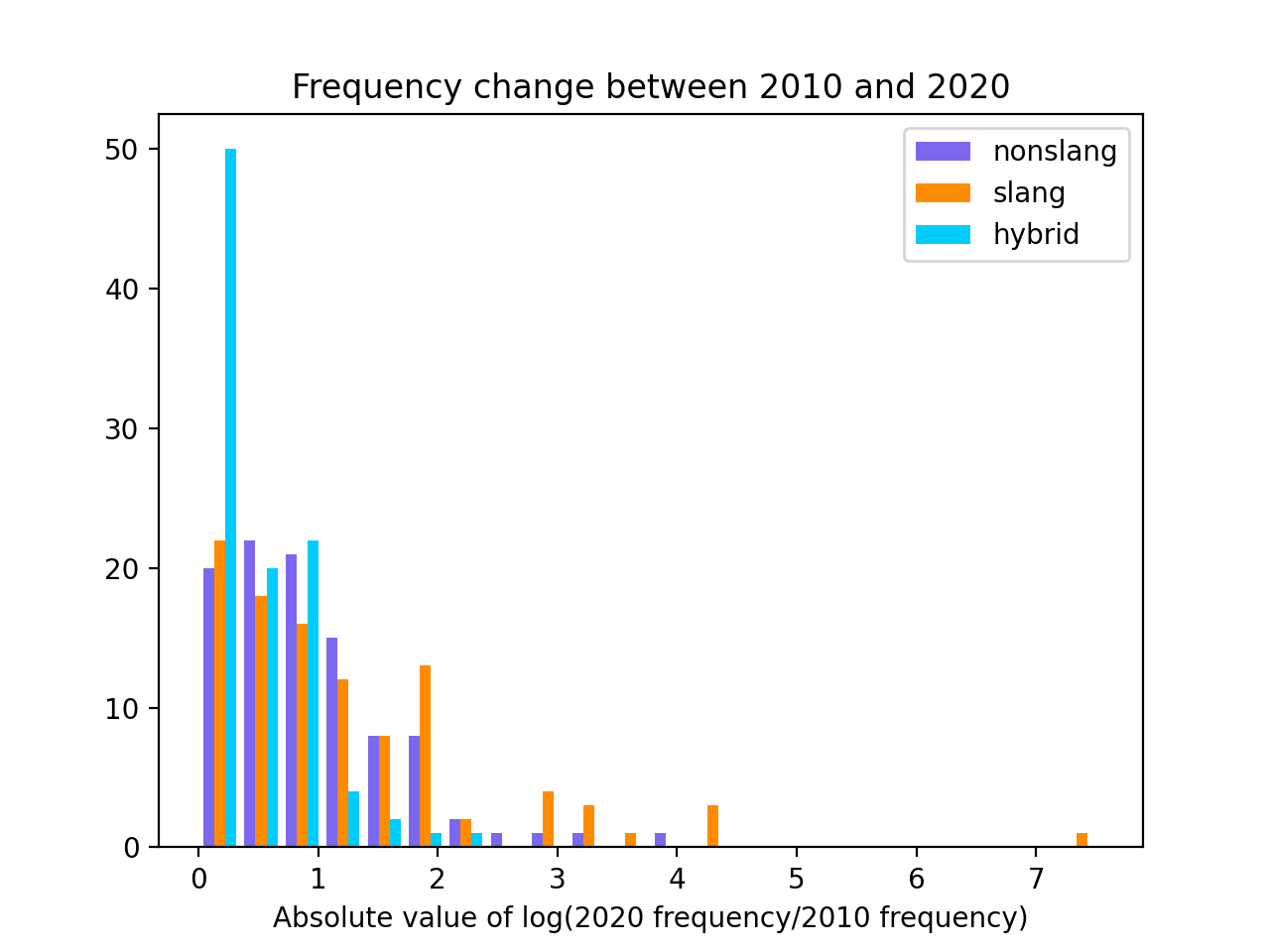}
    \caption{Absolute value of relative difference in frequency between 2020 and 2010, for slang, nonslang and hybrid words, where a larger score corresponds to a larger absolute increase in frequency.}
    \label{fig:hybrid-abs-freq-change}
\end{figure}

In addition, we compare the absolute relative frequency changes as described in \cref{sec:freq_change} across slang, nonslang and hybrid words. The histograms are presented in \cref{fig:hybrid-abs-freq-change}. We observe, respectively, a mean and standard deviation of $1.246$ \& $1.180$ for the slang words, $0.950$ \& $0.724$ for the nonslang words and $0.482$ \& $0.402$ for the hybrid words. The difference in mean is significant between the slang and nonslang words ($p<0.05$), indicating that slang words have undergone a larger absolute change in frequency. Furthermore, we note a highly significant difference ($p<0.001$) in the mean of the hybrid words compared to both the slang and nonslang word means. 

\begin{figure}[t]
    \centering
    \includegraphics[width=0.48\textwidth]{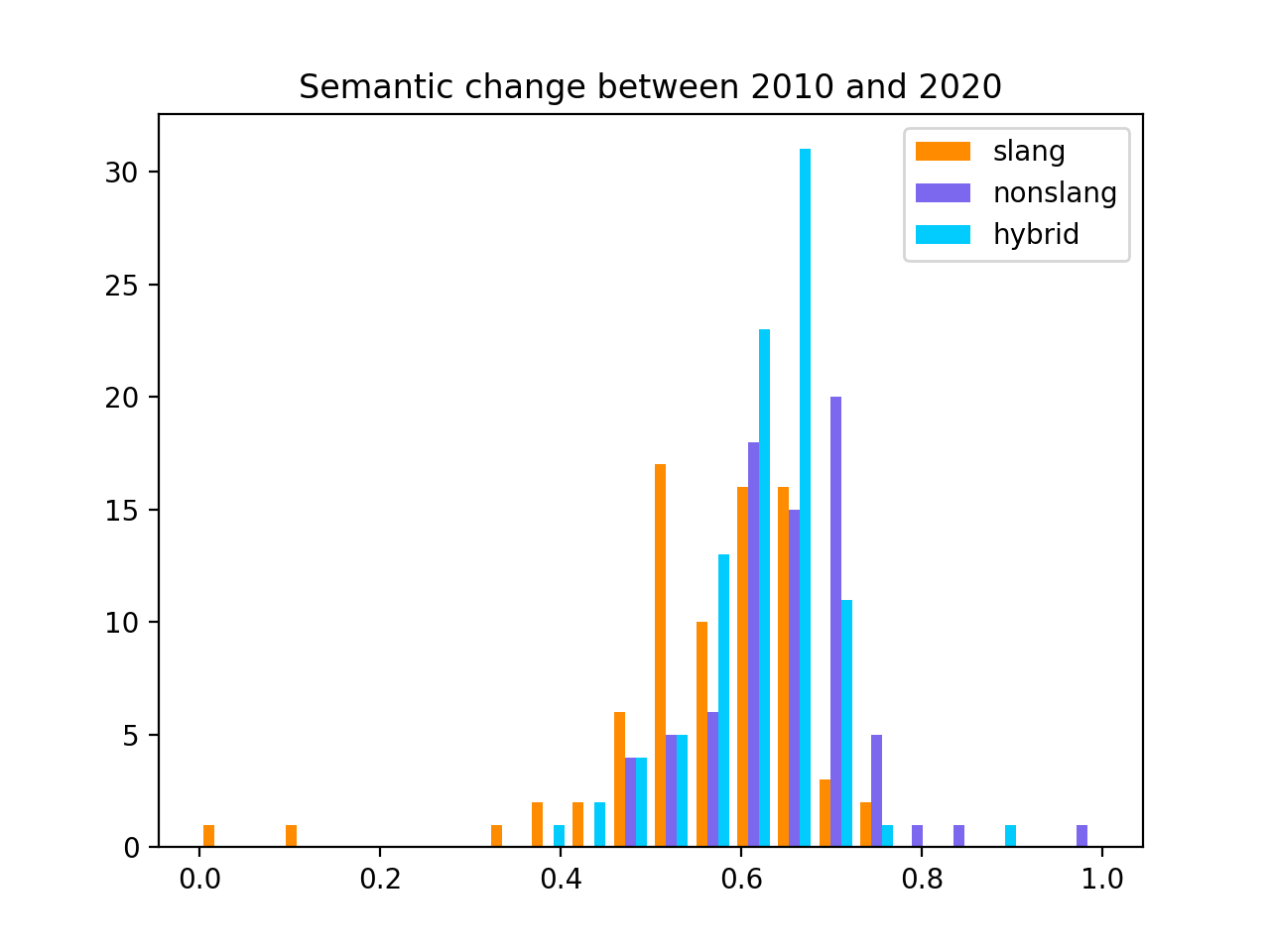}
    \caption{Difference in semantic change score between 2010 and 2020 for slang, nonslang and hybrid words, where a larger score corresponds to a more pronounced semantic change.}
    \label{fig:hybrid-semantic-change}
\end{figure}

We compare the normalized semantic change scores between the slang, nonslang and hybrid words. Histograms over the semantic change scores are shown in \cref{fig:hybrid-semantic-change}. We observe that the distribution over hybrid change scores seem again to be centered between the slang and nonslang distributions, with mean $0.621\pm0.073$. According to a permutation text, there is a significant difference in semantic change both between hybrid and slang words ($p<0.001$) and between hybrid and nonslang words ($p<0.05$). 

\newpage
\section{Appendix -- Causal Analysis}
\subsection{Preliminary on Constraint-based Causal Discovery}
\label{appendix-causal-pre}

\paragraph{Assumptions}The constraint-based causal discovery algorithms make use of two main assumptions, namely the global Markov assumption and the faithfulness assumption. The global Markov property \cite{peters2017elements} holds if all d-separations (defined below) encoded in the causal graph imply conditional independencies in the distribution over the variables contained in the graph. More formally, for a graph $G=(V,E)$ and distribution $\mathbb P$ over the variables $\mathbf X_V$ it holds that for any disjoint subsets $A, B$ and $C$ of $V$
 \[\mathbf X_A \perp_d \mathbf X_B  | \mathbf X_C, \quad \text{in}\; G\]
 \[\Rightarrow \mathbf X_A \indep \mathbf X_B  | \mathbf X_C, \quad \text{in}\;\mathbb P\]
The faithfulness assumption states the converse of the global Markov assumption: All conditional independencies in the distribution are encoded by d-separations in the graph. 

\paragraph{d-separation}Two nodes $A,B\in V$ are said to be \textit{d-separated} \cite{greiger_verma_pearl_bn_indep}, by a set of nodes $Z\subset V$ if for all paths between $A$ and $B$, at least one of the following holds:
\begin{itemize}
    \item The path contains a directed chain ${A\:\cdot\cdot\cdot \rightarrow C \rightarrow \cdot\cdot\cdot\: B}$ or ${A \:\cdot\cdot\cdot\leftarrow C \leftarrow \cdot\cdot\cdot\: B}$ such that ${C\in Z}$
    \item The path contains a fork ${A\:\cdot\cdot\cdot\leftarrow C \rightarrow \cdot\cdot\cdot\: B}$ such that ${C\in Z}$
    \item The path contains a collider ${A\: \cdot\cdot\cdot\rightarrow C \leftarrow \cdot\cdot\cdot\: B}$ such that ${C \notin Z}$ or ${C'\notin Z\quad \forall C'\in desc(C)}$ (i.e., neither $C$ nor any of its descendants is in $Z$)
\end{itemize}
We would then denote $X_A \perp_d X_B | X_Z$. 

\paragraph{Markov Equivalence} Constraint-based algorithms use conditional independence tests 
in order to identify a \textit{Markov equivalence class} of DAGs.
Two DAGs are defined to be Markov equivalent if they have the same skeleton (edges omitting direction) and v-structures. The three vertices $A, B$ and $C$ form a v-structure if $A\rightarrow B \leftarrow C$ and $A$ and $C$ are not directly connected by an edge. Alternatively, two DAGs are Markov equivalent if they describe the same set of d-separation relationships. A Markov equivalence class is the set of all Markov equivalent DAGs.

\paragraph{PC Algorithm} One common constraint-based algorithm is the PC algorithm \cite{spirtes2000causation}. Starting with a full DAG, it eliminates an edge between adjacent vertices $i$ and $j$ if $X_i$ and $X_j$ are conditionally independent given some subset of the remaining variables. This process, including the conditional independence tests, is conducted iteratively starting from a conditioning set of size $k=0$ 
to $k=|V|-2$. In addition to the global Markov and faithfulness assumptions, the PC algorithm also assumes causal sufficiency, namely the absence of unobserved confounders. With these assumptions satisfied and access to correct conditional independence relations, the PC algorithm is guaranteed to be sound, complete and uniformly consistent \cite{kalisch_pc_consistent}. 

\paragraph{PC-stable}PC-stable is an order-independent extension with the same guarantees as the original \cite{colombo_maathuis_14}. 

\subsection{Diagnostic Plots}
\label{appendix-diagnostic}
\begin{figure}
    \centering
    \includegraphics[width=0.5\textwidth]{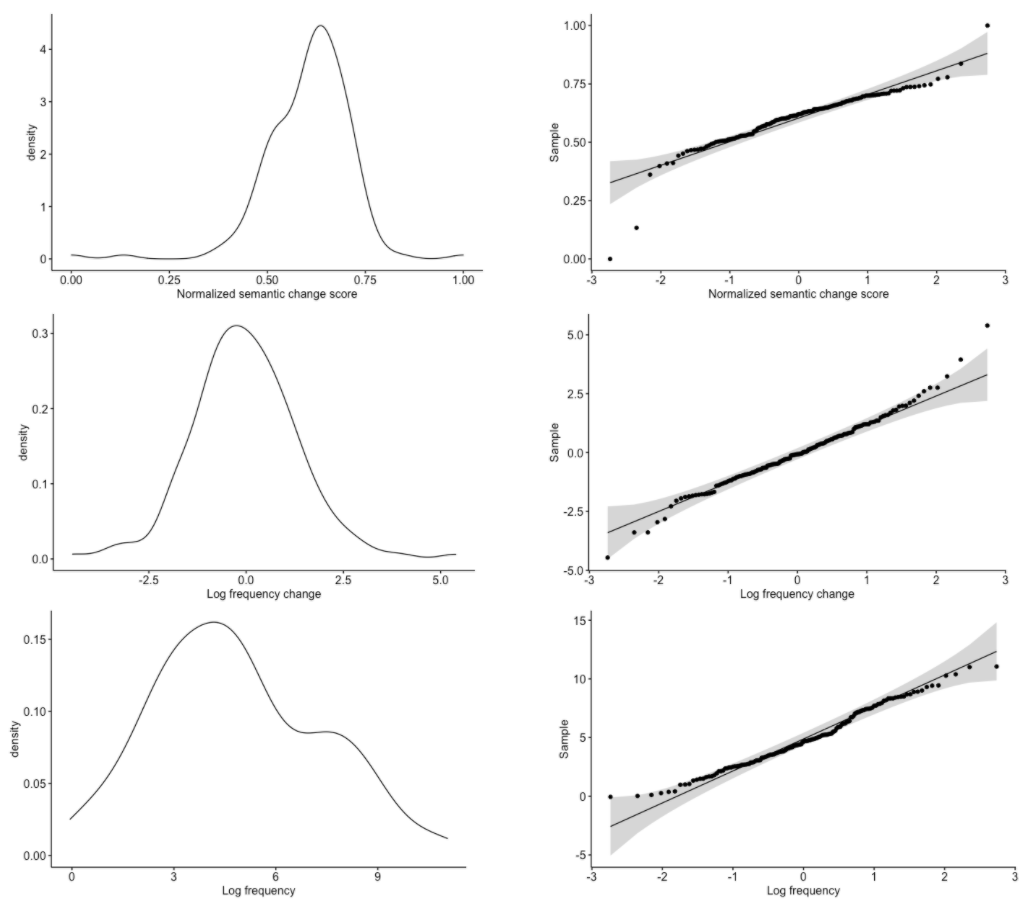}
    \caption{Diagnostic plots for continuous variables, displaying approximate Gaussian shape.}
    \label{fig:diagnostic}
\end{figure}

In \cref{fig:diagnostic} we present the density and Q-Q plots for semantic change score, log of word frequency and log of frequency change.

\subsection{Sensitivity Analysis on Polysemy}
\label{appendix-causal-structure}
\begin{figure}[t]
    \centering
    \includegraphics[width=0.48\textwidth]{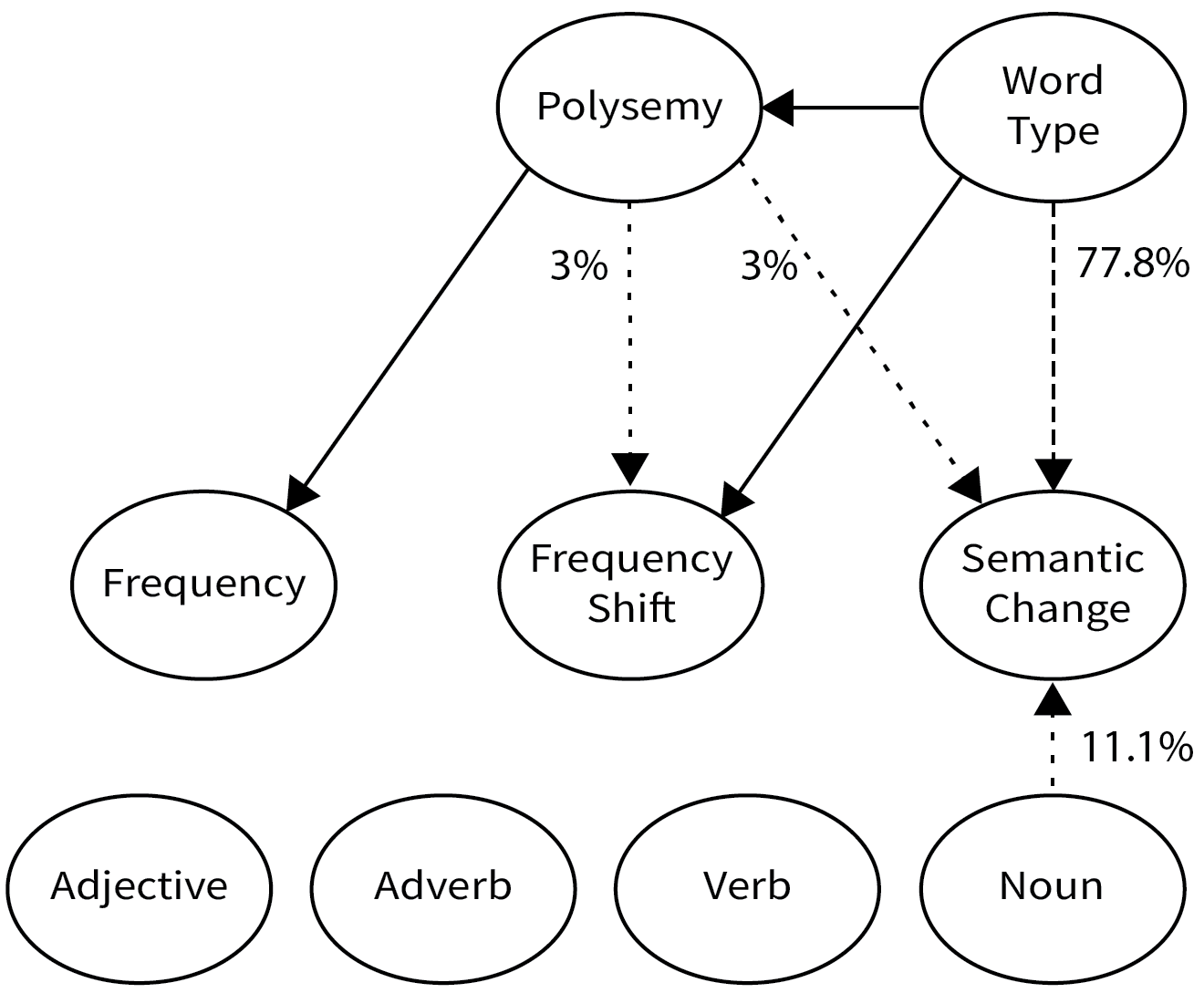}
    \caption{DAG of causal relationships, with the percentage of experiments that found each edge, across different configurations of $\alpha$ and different categorizations of the polysemy score. Solid edges appeared in 100\% of the output graphs.}
    \label{fig:causal_graph_probs}
\end{figure}

Polysemy is a discrete variable which we treat as an ordered factor in the analysis by splitting it into categories. Since polysmey can be plausibly categorized in different ways, we experiment with 9 different categorizations of it and examine the stability of the resulting graphs. For each categorization, we run PC-stable with the three significance levels $\alpha \in \{0.05,0.03,0.01\}$. 
In \cref{fig:causal_graph_probs} we present the results of this sensitivity analysis. 
We see that the edges between word type and polysemy, from word type to frequency change,  as well as the edge from polysemy to frequency, are apparent in all of the configurations. 
The edge from word type to semantic change is apparent in 
21/27 (77.8\%) of the configurations. We also observe a few edges very rarely, and therefore label them as noise and do not take them into account for the causal analysis. These consist of an edge from the POS \textit{Noun} to semantic change in 3/27 (11.1\%) of the configurations, and edges from polysemy to frequency shift and from polysemy to semantic change each apparent in 1/27 (3.7\%) of the configurations. 

By inferring the causal graph from a set of categorizations, we make up for the possible noise in the polysemy variable and ensure that the graph is not sensitive to small variations in the words' polysemy scores. 

\subsection{Causal Inference}
\label{appendix-causal-inference}
Given the causal DAG in \cref{fig:causal-graph}, we derive the expression for the average causal effect of word type on semantic change. Define the following random variables: $T=$ word type, $X=$ polysemy, $Y=$ frequency, $Z=$ frequency shift and $S=$ semantic change, with respective probability mass functions $P_T$ \& $P_X$ and probability density functions $f_Y$, $f_Z$ \& $f_S$.

Note that the possible values for $T$ lie in $\{\text{slang}, \text{nonslang}\}$. By the truncated factorization for the connected component of the causal DAG (i.e., excluding POS), we have that
\[\mathbb P(s,t,x,y,z|do(T=t'))=\]
\[f_{Y|X}(y|x)f_{Z|T}(z|t)f_{S|T}(s|t)P_{X|T}(x|t)\mathbbm 1_{\{t=t'\}}\]
Marginalizing over $T$, we get
\[\mathbb P(s,x,y,z|do(T=t'))=\]
\[=f_{Y|X}(y|x)f_{Z|T}(z|t')f_{S|T}(s|t')P_{X|T}(x|t')\]
Next, marginalize over the continuous random variables $Y$ and $Z$ to get
\[\mathbb P(s,x|do(T=t'))=\]
\[\int_y\int_zf_{Y|X}(y|x)f_{Z|T}(z|t')f_{S|T}(s|t')P_{X|T}(x|t')dzdy=\]
\[\int_yf_{Y|X}(y|x)f_{S|T}(s|t')P_{X|T}(x|t')\underbrace{\left(\int_zf_{Z|T}(z|t')dz\right)}_{=1}dy=\]
\[f_{S|T}(s|t')P_{X|T}(x|t')\underbrace{\int_yf_{Y|X}(y|x)dy}_{=1}=\]
\[f_{S|T}(s|t')P_{X|T}(x|t')\]

Finally
\[\mathbb P(s|do(T=t'))=\]
\[\sum_{x}f_{S|T}(s|t')P_{X|T}(x|t')=f_{S|T}(s|t')\]

Taking the expectation, we get
\[\mathbb E[S|do(T=t')]=\mathbb E_{S|T}[S|t']\]

\section{Appendix -- Selected Words}
\label{appendix-selected-words}
In \cref{tab:wordlist} we list all the slang and nonslang words used in this study. 
\begin{center}
\begin{tabular}{ c c c }
 \toprule
 \textbf{Slang} & \textbf{Nonslang} & \textbf{Hybrid} \\ \midrule  
 a-list & admitting & annihilated \\
badass & adulterous & balling\\
blankie & agenda & bastard\\
bling & allotted & beef\\
blowjob & anticlockwise & bloody\\
blumpkin & avoiders & bomb\\
bonehead & awesome & book\\
bro & banzai & bookmark\\
bromance & bright & booty\\
bumfuck & butane & bounce\\
bupkis & calorie & bowl\\
chillax & chug & brains\\
chones & committeeman & candle\\
colitas & competencies & chicken\\
compo & contenders & classic\\
conniption & conventionally & crock\\
crappy & copyediting & decompress\\
dang & deathblow & dim\\
dis & decomposition & dirt\\
dogg & despoil & dose\\
duckface & didot & down\\
dudette & doubleheader & egg\\
fanboy & echo & eye\\
fap & enhancements & fat\\
gangsta & epilator & fence\\
glitterati & estimated & fire\\
gorp & fiddled & fluffer\\
gotsta & galavant & foxy\\
gunt & glutton & freckle\\
hasbian & greeting & fruitcake\\
horribad & grisly & gag\\
jabroni & groans & ghost\\
jalopy & haircut & gig\\
jerkwad & heaviest & gnarly\\
lame-o & humblest & god\\
lemme & ignites & gridlock\\
lowkey & inclusive & grip\\
mcdreamy & intimidator & grub\\
meme & jugglers & gumby\\
mosey & jute & hanger\\
motherfucking & lawlessness & head\\
mozzie & legalist & hell\\
netizen & milepost & hitter\\
nuker & mistreatment & item\\
pedo & moldovan & jammed\\
peeps & morphology & jill\\
plastered & mushroom & jock\\
poopy & nonskid & kick\\
preemie & outlawing & kosher\\
pregos & pantsuit & locks\\
prettyful & peppy & mad\\
rapey & performative & mine\\
\end{tabular}
\label{tab:wordlist}
\end{center}

\begin{center}
    \begin{tabular}{ c c c }
 \toprule
 \textbf{Slang} & \textbf{Nonslang} & \textbf{Hybrid}\\ \midrule  
rehab & postural &  money\\
relly & protocol &  move\\
roofie & repentant & mule\\
roshambo & rump & pecker\\
sesh & sabertooth & peckish\\
shart & sailor & peeper\\
shiesty & scallywag & pig\\
shtick & scheme & pinch\\
sicc & sculptured & plums\\
sinse & scummiest & postal\\
skeevy & shield & rad\\
skyrocket & shylock & ratchet\\
slore & snug & roadkill\\
snitch & squall & sausage\\
soused & steeple & scissor\\
spam & strap & scoot\\
spec & superabundance & scream\\
spec-ops & sympathizer & screaming\\
sucky & telogen & smoked\\
tenner & terrifies & sneak\\
thingamabob & they & split\\
trisexual & trampolining & squawk\\
tweeker & underpainting & stat\\
twit & underrated & stew\\
whadja & unicorn & streak\\
workaround & unlike & styling\\
wut & unmatched & swap\\
zooted & upgrade & thick\\
 & vanadium & thirsty\\
  &  & threads \\
  &  & tool \\
  &  & toots \\
  &  & tweaker \\
  &  & walk \\
  &  & walkie \\
  &  & whippet \\
  &  & windy \\
  &  & wrecked \\
  &  & zombie \\
  &  & zounds \\
    \bottomrule
\end{tabular}
\end{center}
\end{document}